\documentclass[10pt,journal,compsoc]{IEEEtran}
\usepackage{multicol}
\pdfoutput=1
\usepackage{times}
\usepackage{epsfig}
\usepackage{graphicx} 
\usepackage{amsmath}
\usepackage{amssymb}

\usepackage{color}

\usepackage{balance} 
\usepackage{mathtools}
\usepackage{subcaption} 
\usepackage{textcomp} 
\usepackage{siunitx}
\usepackage{booktabs} 

\usepackage{float}

\floatstyle{ruled}
\newfloat{algorithm}{tbp}{loa}
\providecommand{\algorithmname}{Algorithm}
\floatname{algorithm}{\protect\algorithmname}

\newif\ifshorter
\shorterfalse

\global\long\def\bu{\mathbf{u}}

\global\long\def\bxi{\boldsymbol{\xi}}

\global\long\def\thr{C_{\text{th}}} 
\global\long\def\Iref{I^r} 
\global\long\def\Nref{N_r} 
\global\long\def\pose{\bxi}
\global\long\def\obs{o} 
\global\long\def\state{s} 
\global\long\def\natparam{\eta} 
\global\long\def\basemeas{h} 
\global\long\def\suffStat{T} 
\global\long\def\KLdiv{D_{\mathrm{KL}}} 
\global\long\def\Mo{\bar{M}_k} 
\global\long\def\JacMeas{J_k} 
\global\long\def\inlierProb{\pi_{m}} 
\global\long\def\stdThr{\sigma_{m}} 
\global\long\def\varThr{\stdThr^2} 
\global\long\def\cN{\mathcal{N}} 
\global\long\def\cU{\mathcal{U}} 

\global\long\def\log{\ln} 

\usepackage{bbold}
\global\long\def\identity{\mathbb{1}} 

\usepackage[pagebackref=true,breaklinks=true,letterpaper=true,colorlinks,bookmarks=true]{hyperref}

\usepackage{xspace}
\newcommand*{\eg}{e.g.,\@\xspace}
\newcommand*{\ie}{i.e.,\@\xspace}

\ifCLASSOPTIONcompsoc
  \usepackage[nocompress]{cite}
\else
  \usepackage{cite}
\fi
\ifCLASSINFOpdf
\else
\fi
\begin{document}
%
\title{Event-based, 6-DOF Camera Tracking from Photometric Depth Maps}
%
%
%
%

\author{Guillermo~Gallego, Jon~E.A.~Lund, Elias~Mueggler, Henri~Rebecq, Tobi~Delbruck, Davide~Scaramuzza
\IEEEcompsocitemizethanks{\IEEEcompsocthanksitem The authors are with the Robotics and Perception Group, affiliated with both the Dept. of Informatics of the University of Zurich and the Dept. of Neuroinformatics of the University of Zurich and ETH Zurich, Switzerland: \url{http://rpg.ifi.uzh.ch/}.
This research was supported by the National Centre of Competence in Research (NCCR) Robotics, the SNSF-ERC Starting Grant, the Qualcomm Innovation Fellowship, the DARPA FLA program, and the UZH Forschungskredit.
}
}

%
%

\markboth{IEEE Transactions on Pattern Analysis and Machine Intelligence,~Vol.~?, No.~?, 2017}%
{Gallego \MakeLowercase{\textit{et al.}}: Event-based, 6-DOF Camera Tracking from Photometric Depth Maps}
%



\IEEEtitleabstractindextext{%
\begin{abstract}
Event cameras are bio-inspired vision sensors that output pixel-level brightness changes instead of standard intensity frames.
These cameras do not suffer from motion blur and have a very high dynamic range, which enables them to provide reliable 
visual information during high-speed motions or in scenes characterized by high dynamic range.
These features, along with a very low power consumption, make event cameras an ideal complement to standard cameras for VR/AR and video game applications.
With these applications in mind, this paper tackles the problem of accurate, low-latency tracking of an event camera from an 
existing photometric depth map (i.e., intensity plus depth information) built via classic dense reconstruction pipelines.
Our approach tracks the \mbox{6-DOF} pose of the event camera upon the arrival of each event, thus virtually eliminating latency.
We successfully evaluate the method in both indoor and outdoor scenes and
show that---because of the technological advantages of the event camera---our pipeline works in scenes characterized by high-speed motion, which are still unaccessible to standard cameras.
\end{abstract}

}

\maketitle

\IEEEdisplaynontitleabstractindextext

%
\IEEEpeerreviewmaketitle

\IEEEraisesectionheading{\section*{Supplementary Material}}

Video of the experiments: \href{https://youtu.be/iZZ77F-hwzs}{https://youtu.be/iZZ77F-hwzs}.


\section{Introduction}
\IEEEPARstart{T}{he} 
task of estimating a sensor's ego-motion has important applications in various fields, such as augmented/virtual reality (AR/VR), video gaming, and autonomous mobile robotics.
In recent years, great progress has been achieved using visual information to fulfill such a task \cite{MurArtal15tro,Engel14eccv,Forster17troSVO}.
However, due to some well-known limitations of traditional cameras (motion blur and low dynamic-range), current visual odometry 
pipelines still struggle to cope with high-speed motions or high dynamic range scenarios.
Novel types of sensors, called event cameras~{\protect\cite[p.77]{belbachir09book}}, offer great potential to overcome these issues.

\begin{figure}[t]
\begin{flushleft}
\includegraphics[trim={3.5cm 0 1.5cm 1.1cm},clip,width=0.76\columnwidth]{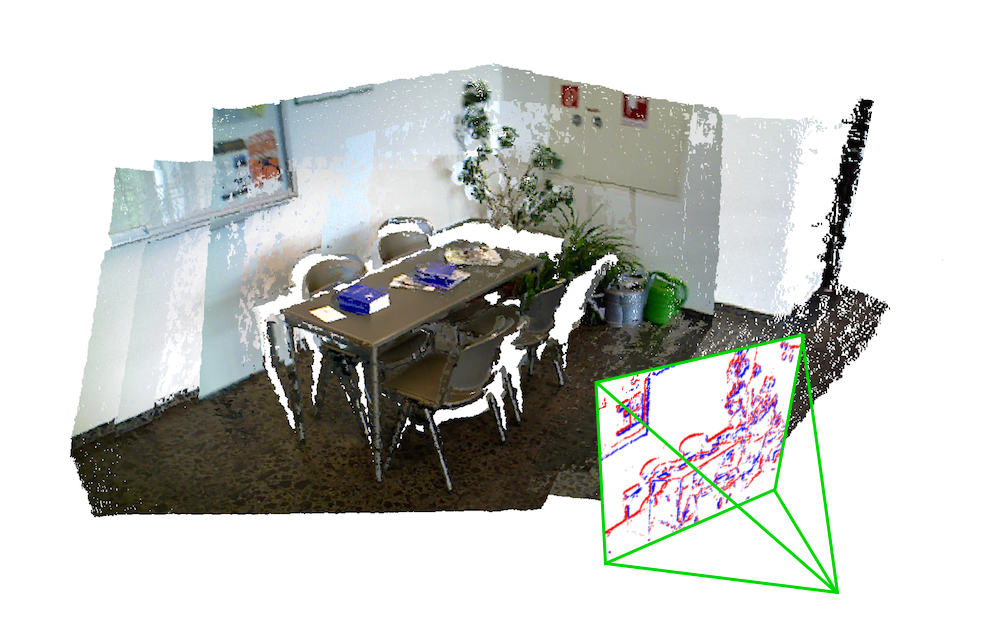}
\llap{\makebox[0.7cm][l]{\raisebox{-0.5cm}{\frame{\includegraphics[height=1.8cm]{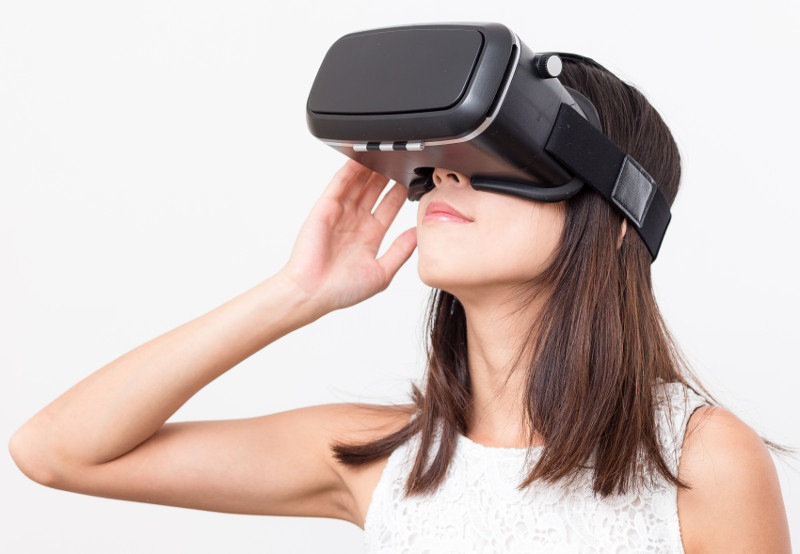}}}}}
\end{flushleft}
   \caption{Sample application: \mbox{6-DOF} tracking in AR/VR (Augmented or Virtual Reality) scenarios. 
   The pose of the event camera (rigidly attached to a hand or head tracker) is tracked from a previously built photometric depth map (RGB-D) of the scene.
   Positive and negative events are represented in blue and red, respectively, on the image plane of the event camera.}
\label{fig:EyCatcherNew}
\end{figure}

Unlike standard cameras, which transmit intensity frames at a fixed framerate, event cameras, such as the Dynamic Vision Sensor (DVS)~\cite{Lichtsteiner06isscc}, 
only transmit \emph{changes of intensity}. Specifically, they transmit per-pixel intensity changes at the time they occur, 
in the form of a set of asynchronous \emph{events}, where each event carries the space-time coordinates of the brightness change (with microsecond resolution) and its sign.

Event cameras have numerous advantages over standard cameras: a latency in the order of microseconds, a very high dynamic range (140 dB compared to 60 dB of standard cameras), 
and very low power consumption (10 mW vs 1.5 W of standard cameras).
Most importantly, since all pixels capture light \emph{independently}, such sensors do not suffer from motion blur.

It has been shown that event cameras transmit, in principle, all the information needed to reconstruct a full video stream \cite{Cook11ijcnn,Bardow16cvpr,Reinbacher16bmvc,Rebecq17ral}, 
which clearly points out that an event camera alone is sufficient to perform \mbox{6-DOF} state estimation and 3D reconstruction.
Indeed, this has been recently shown in~\cite{Kim16eccv,Rebecq17ral}. However, currently the quality of the 3D map built using event cameras does not achieve the same level of detail and accuracy as that of standard cameras.

Although event cameras have become commercially available only since 
2008~\cite{Lichtsteiner08ssc}, the recent body of literature on these new sensors\footnote{\url{https://github.com/uzh-rpg/event-based_vision_resources}} 
as well as the recent plans for mass production claimed by 
companies, such as Samsung and Chronocam\footnote{\url{http://rpg.ifi.uzh.ch/ICRA17_event_vision_workshop.html}}, 
highlight that there is a big commercial interest in exploiting these new vision sensors 
as an ideal complement to standard cameras for mobile robotics, VR/AR, and video game applications.

Motivated by these recent developments, this paper tackles the problem of 
tracking the \mbox{6-DOF} motion of an event camera from an RGB-D (i.e., photometric depth) map that has been previously built via a
traditional, dense reconstruction pipeline using standard cameras or RGB-D sensors (cf. Fig.~\ref{fig:EyCatcherNew}).
This problem is particularly important in both AR/VR and video game applications,
where low-power consumption and robustness to high-speed motion are still unsolved. In these applications, we envision that the user would first use a standard sensor to build a high resolution and high quality map of the room, and then the hand and head trackers would take advantage of an event camera to achieve robustness to high-speed motion and low-power consumption.

\begin{figure}[t]
\begin{center}
\frame{\includegraphics[height=1.22in]{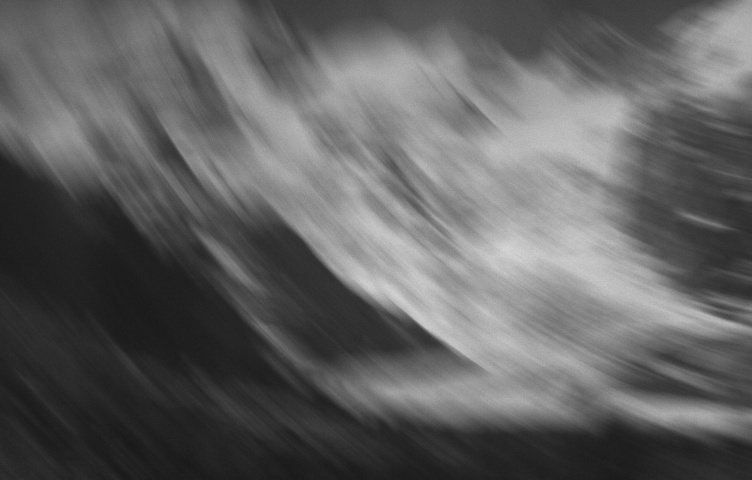}}~
\frame{\includegraphics[height=1.22in]{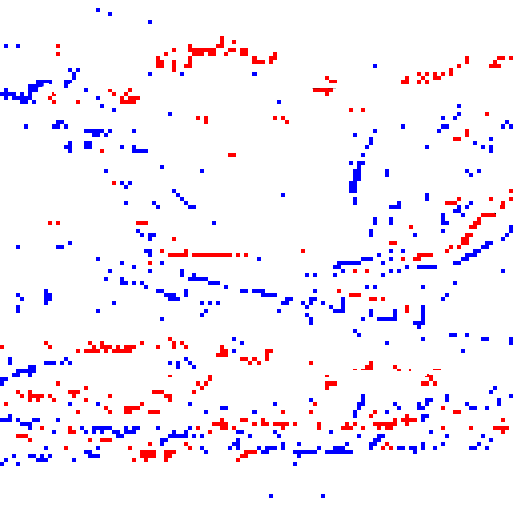}}\\[2mm]
   \includegraphics[width=1.0\linewidth]{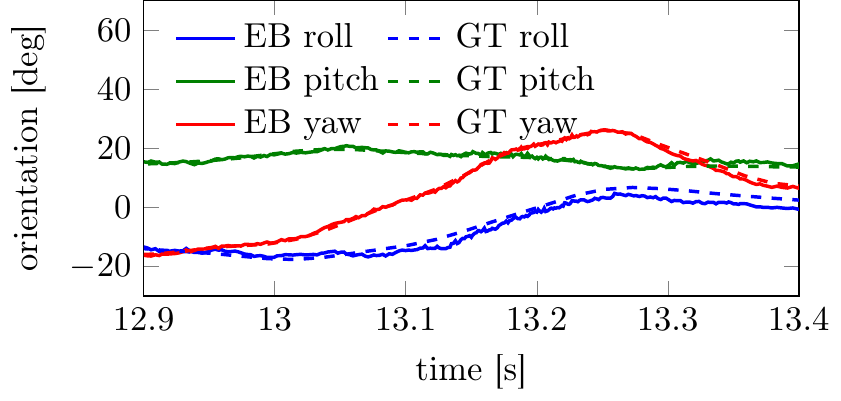}
\end{center}
   \caption{High-speed motion sequence. Top left: image from a standard camera, suffering from blur due to high-speed motion.
   Top right: set of asynchronous events from a DVS in an interval of 3~milliseconds, colored according to polarity. 
   Bottom: estimated poses using our event-based (EB) approach, which provides low latency and high temporal resolution updates. 
   Ground truth (GT) poses are also displayed.
	}
\label{fig:EyCatcher}
\end{figure}

The challenges we address in this paper are two: 
$i$) event-based \mbox{6-DOF} pose tracking from an existing photometric depth map;
$ii$) tracking the pose during very fast motions 
(still unaccessible to standard cameras because of motion blur), as shown in~\figurename~\ref{fig:EyCatcher}. 
We show that we can track the \mbox{6-DOF} motion of the event camera with comparable accuracy as that of standard cameras and also during high-speed motion.

Our method is based on Bayesian filtering theory and has three key contributions in the way that the events are processed: 
$i$) 
event-based pose update, meaning that the \mbox{6-DOF} pose estimate is updated every time an event is generated, at \emph{micro}second time resolution,
$ii$) 
the design of a sensor likelihood function using a mixture model that takes into account both the event generation process and the presence of noise and outliers (Section~\ref{sec:MeasurementModel}), 
and $iii$) 
the approximation of the posterior distribution of the system by a tractable distribution in the exponential family, which is obtained by minimizing the Kullback-Leibler divergence (Section~\ref{sec:PosteriorApprox}).
The result is a filter adapted to the asynchronous nature of the event camera, which also incorporates an outlier detector that weighs measurements according to their confidence for improved robustness of the pose estimation.
The approximation of the posterior distribution allows us to obtain a closed-form solution to the filter update equations and has the benefit of being computationally efficient. 
Our method can handle arbitrary, \mbox{6-DOF}, high-speed motions of the event camera in natural scenes.

The paper is organized as follows: 
Section~\ref{sec:RelatedWorkEgoMotion} reviews related literature on event-based ego-motion estimation. 
Section~\ref{sec:DVSdescription} describes the operating principle of event cameras.
Our proposed event-based, probabilistic approach is described in Section~\ref{sec:Methodology}, 
and it is empirically evaluated on natural scenes in Section~\ref{sec:Evaluation}. 
Conclusions are highlighted in Section~\ref{sec:Conclusion}.

\section{Related work on Event-based Ego-Motion Estimation}
\label{sec:RelatedWorkEgoMotion}

The first work on pose tracking with a DVS was presented in~\cite{Weikersdorfer12robio}. 
The system design, however, was limited to slow planar motions (i.e., 3 DOF) and planar scenes parallel to the plane of motion consisting of artificial B\&W line patterns.
The particle filter pose tracker was extended to 3D in~\cite{Weikersdorfer14icra}, 
where it was used in combination with an external RGB-D sensor (depth estimation) to build a SLAM system.
However, a depth sensor introduces the same bottlenecks that exist in standard frame-based systems: 
depth measurements are outdated for very fast motions, and the depth sensor is still susceptible to motion blur.

In our previous work~\cite{Censi14icra}, a standard grayscale camera was attached to a DVS to estimate the small displacement between the current event and the previous frame of the standard camera.
The system was developed for planar motion and artificial B\&W striped background. 
This was due to the sensor likelihood being proportional to the magnitude of the image gradient, 
thus favoring scenes where large brightness gradients are the source of most of the event data.
Because of the reliance on a standard camera, the system was again susceptible to motion blur and therefore limited to slow motions.

An event-based algorithm to track the \mbox{6-DOF} pose of a DVS alone and during very high-speed motion was presented in~\cite{Mueggler14iros}. 
However, the method was developed specifically for artificial, B\&W line-based maps. 
Indeed, the system worked by minimizing the point-to-line reprojection error.

Estimation of the 3D orientation of an event camera was presented in~\cite{Cook11ijcnn,Kim14bmvc,Gallego17ral,Reinbacher17iccp}. 
However, such systems are restricted to rotational motions, and, thus, do not account for translation and depth.

Contrarily to all previous works, the approach we present in this paper tackles full \mbox{6-DOF} motions, does not rely on external sensors, can handle arbitrary fast motions, and is not restricted to specific texture or artificial scenes.

Other pose tracking approaches have been published as part of systems that address the event-based 3D SLAM problem. 
\cite{Kim16eccv} proposes a system with three interleaved probabilistic filters to perform pose tracking as well as depth and intensity estimation. 
The system is computationally intensive, requiring a GPU for real-time operation.
The parallel tracking-and-mapping system in \cite{Rebecq17ral} follows a geometric, semi-dense approach.
The pose tracker is based on edge-map alignment and the scene depth is estimated without intensity reconstruction, 
thus allowing the system to run in real-time on the CPU.
More recently, visual inertial odometry systems based on event cameras have also been proposed, which rely on point features \cite{Mueggler17tro,Zhu17cvpr,Rebecq17bmvc}.

\section{Event Cameras}
\label{sec:DVSdescription}

\begin{figure}[t!]
    \centering
    \begin{subfigure}[t]{\linewidth}
        \centering
        \includegraphics[width=0.8\linewidth]{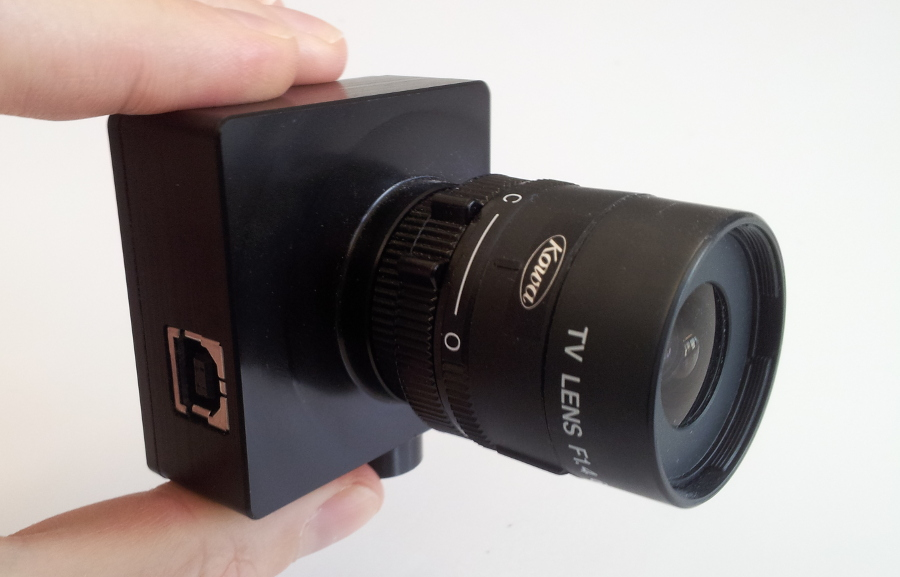}\\
        \caption{The Dynamic Vision Sensor (DVS) from iniLabs.}
        \label{fig:DVSDevice}
    \end{subfigure} \\[1mm]
    \begin{subfigure}[t]{\linewidth}
        \centering
        \includegraphics[width=0.9\linewidth]{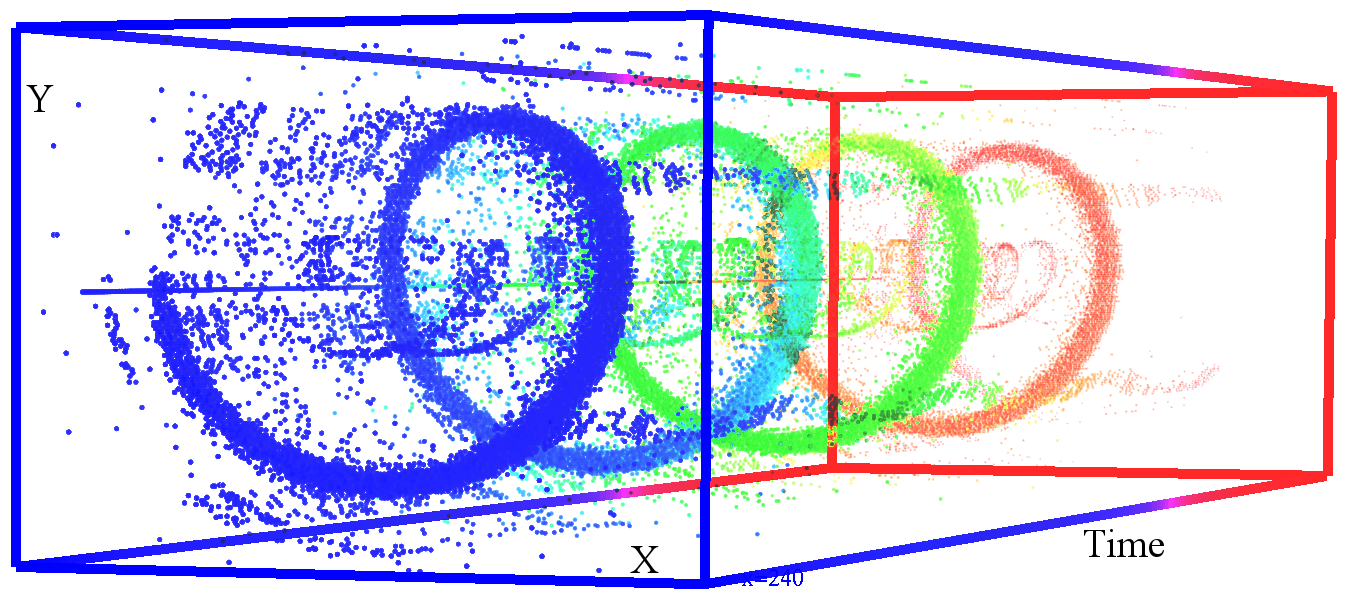}
        \caption{Visualization of the output of a DVS (event stream) while viewing a rotating scene, which generates a spiral-like structure in space-time. 
   Events are represented by colored dots, from red (far in time) to blue (close in time).
   Event polarity is not displayed. 
   Noise is visible by isolated points.}
        \label{fig:DVSSpaceTime}
    \end{subfigure}
    \caption{An event camera and its output.}
    \label{fig:DVSstream}
\end{figure}

Event-based vision constitutes a paradigm shift from conventional (\eg frame-based) vision.
In standard cameras, pixels are acquired and transmitted simultaneously at fixed rates; 
this is the case of both global-shutter or rolling-shutter sensors.
Such sensors provide little information about the scene in the ``blind time'' between consecutive images.
Instead, event-based cameras such as the DVS~\cite{Lichtsteiner08ssc} (\figurename~\ref{fig:DVSDevice}) have independent pixels that respond asynchronously to relative contrast changes.
If $I(\bu,t)$ is the intensity sensed at a pixel $\bu=(x,y)^\top$ of the DVS, 
an event is generated if the temporal visual contrast (in log scale) exceeds a nominal threshold $\thr$:
\begin{equation}
\label{eq:EventTrigCondition}
\Delta\ln I \coloneqq \ln I(\bu,t) - \ln I(\bu,t-\Delta t) \gtrless \thr,
\end{equation}
where $\Delta t$ is the time since the last event was generated at the same pixel.
Different thresholds may be specified for the cases of contrast increase ($\thr^{+}$) or decrease $(\thr^{-})$.
An event $e=(x,y,t,p)$ conveys the spatio-temporal coordinates and sign (\ie polarity) of the brightness change, 
with $p=+1$ (ON-event: $\Delta \ln I>\thr^{+}$) or $p=-1$ (OFF-event: $\Delta \ln I<\thr^{-}$).
Events are time-stamped with microsecond resolution and transmitted asynchronously when they occur, with very low latency.
A sample output of the DVS is shown in \figurename~\ref{fig:DVSSpaceTime}.
Another advantage of the DVS is its very high dynamic range (\SI{140}{\dB}), which notably exceeds the \SI{60}{\dB} of high-quality, conventional frame-based cameras.
This is a consequence of events triggering on log-intensity changes~\eqref{eq:EventTrigCondition} instead of absolute intensity.
The spatial resolution of the DVS is $128\times 128$ pixels, but newer sensors, such as the Dynamic and Active-pixel VIsion Sensor (DAVIS)~\cite{Brandli14ssc}, the color DAVIS (C-DAVIS)~\cite{Li15iiws}, and the Samsung DVS~\cite{Son17isscc} have higher resolution ($640\times 480$ pixels), thus overcoming current limitations.

\section{Probabilistic approach}
\label{sec:Methodology}

Consider an event camera moving in a known static scene. 
The map of the scene is described by a sparse set of reference images $\{\Iref_l\}_{l=1}^{\Nref}$, poses $\{\pose^r_l\}_{l=1}^{\Nref}$, and depth map(s).
Suppose that an initial guess of the location of the event camera in the scene is also known.
The problem we face is that of exploiting the information conveyed by the event stream to track the pose of the event camera in the scene.
Our goal is to handle arbitrary \mbox{6-DOF}, high-speed motions of the event camera in realistic (\ie natural) scenes.

We design a robust filter combining the principles of Bayesian estimation, posterior approximation, and exponential family distributions with a sensor model that accounts for outlier observations. 
In addition to tracking the kinematic state of the event camera, the filter also estimates some sensor parameters automatically (\eg event triggering threshold $\thr$) that would otherwise be difficult to tune manually.
\footnote{Today's event-based cameras, such as the DVS~\cite{Lichtsteiner08ssc} or the DAVIS~\cite{Brandli14ssc}, have almost a dozen tuning parameters that are neither independent nor linear.}

The outline of this section is as follows. 
First, the problem is formulated as a marginalized posterior estimation problem in a Bayesian framework. 
Then, the motion model and the measurement model 
(a robust likelihood function that can handle both good events and outliers) are presented. 
Finally, the filter equations that update the parameters of an approximate distribution to the posterior probability distribution are derived.

\subsection{Bayesian Filtering}
\label{sec:BayesianFiltering}

We model the problem as a time-evolving system whose state $s$ consists of the kinematic description of the event camera as well as sensor and inlier/outlier parameters.
More specifically,
\begin{equation}
\label{eq:StateVecGeneric}
\state = (\bxi_c,\bxi_{i},\bxi_{j},\thr,\pi_m,\sigma^2_m)^\top,
\end{equation}
where $\bxi_c$ is the current pose of the sensor (at the time of the event, $t$ in~\eqref{eq:EventTrigCondition}), 
$\bxi_{i}$ and $\bxi_{j}$ are two poses along the sensor's trajectory that are used to interpolate the pose of the last event at the same pixel (time $t-\Delta t$ in ~\eqref{eq:EventTrigCondition}), $\thr$ is the contrast threshold, and $\pi_m$ and $\sigma_m^2$ are the inlier parameters of the sensor model, which is explained in Section~\ref{sec:MixtureLikelihoodFunction}.

\begin{figure}[t!]
\centering
   \includegraphics[width=\linewidth]{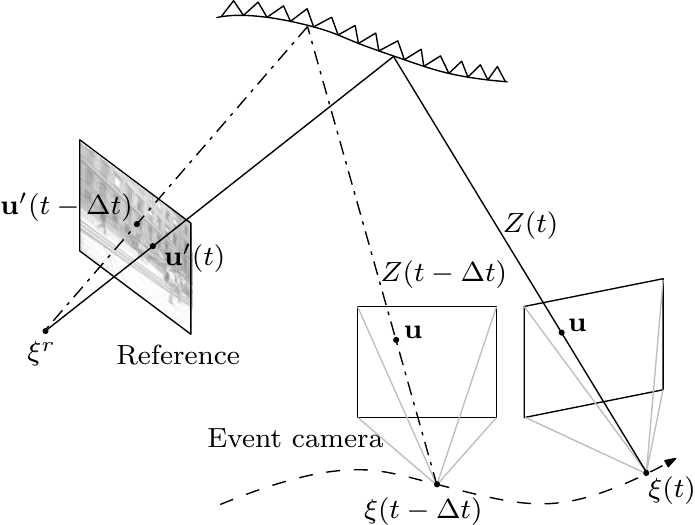}
   \caption{Computation of the contrast (measurement function) by transferring events from the event camera to a reference image. 
   For each event, the predicted contrast~\eqref{eq:ApproxContrastOnRefImage}, $\Delta \ln I$, used in the measurement function~\eqref{eq:Measurementfunction} is computed as the log-intensity difference (as in~\eqref{eq:EventTrigCondition}) at two points on the reference image $I^r$: 
the points~\eqref{eq:transferredPoint} corresponding to the same pixel $\bu$ on the event camera, at times of the event ($t_k$ and $t_k-\Delta t$).
}
\label{fig:TransferEvents}
\end{figure}

Let the state of the system at time $t_k$ be $\state_k$, and let the sequence of all past observations (up to time $t_k$) be $\obs_{1:k}$, where $\obs_k$ is the current observation (\ie the latest event).

Our knowledge of the system state is contained in the posterior probability distribution $p(\state_k|\obs_{1:k})$, also known as \emph{belief}~\cite[p.27]{Thrun05book}, which is the marginalized distribution of the smoothing problem $p(\state_{1:k}|\obs_{1:k})$. 
The Bayes filter recursively estimates the system state from the observations in two steps: prediction and correction.
The correction step updates the posterior by:
\begin{equation}
\label{eq:BayesCorrectionStepMargPosterior}
p(\state_{k}|\obs_{1:k}) \propto p(\obs_{k}|\state_{k}) p(\state_{k}|\obs_{1:k-1}),
\end{equation}
where $p(\obs_{k}|\state_{k})$ is the likelihood function (sensor model)
and we used independence of the events given the state.
The prediction step, defined by
\begin{equation}
\label{eq:BayesPredStepMargPosterior}
p(\state_{k}|\obs_{1:k-1}) = \int p(\state_{k}|\state_{k-1}) p(\state_{k-1}|\obs_{1:k-1}) d\state_{k-1},
\end{equation}
incorporates the motion model $p(\state_{k}|\state_{k-1})$ from $t_{k-1}$ to $t_{k}$.

We incorporate in our state vector not only the current event camera pose $\bxi_c^k$ but also the other relevant poses for contrast calculation 
(poses $\bxi_i^k,\bxi_j^k$ in~\eqref{eq:StateVecGeneric}), 
so that we may use the filter to partially correct errors of already estimated poses.
Past events that are affected by the previous pose are not re-evaluated, 
but future events that reference back to such time will have better previous-pose estimates.

To have a computationally feasible filter, we approximate the posterior~\eqref{eq:BayesCorrectionStepMargPosterior} by a tractable distribution with parameters $\natparam_{k-1}$ that condense the history of events $\obs_{1:k-1}$,
\begin{equation}
\label{eq:approxPosteriorDef}
p(\state_{k}|\obs_{1:k}) \approx q(\state_{k};\natparam_{k}).
\end{equation}
Assuming a motion model with slowly varying zero-mean random diffusion,
so that most updates of the state are due to the events, the recursion on the approximate posterior becomes, combining~\eqref{eq:BayesCorrectionStepMargPosterior}-\eqref{eq:approxPosteriorDef},
\begin{equation}
\label{eq:BayesCorrectionStepMargPostApprox}
q(\state_{k};\natparam_{k}) \approx C\, p(\obs_{k}|\state_{k}) q(\state_{k};\natparam_{k-1})
\end{equation} 
for some normalizing constant $C$.
The approximate posterior~$q$ is computed by minimization of the Kullback-Leibler (KL) divergence 
between both sides of~\eqref{eq:BayesCorrectionStepMargPostApprox}.
As tractable distribution we choose one in the exponential family because they are very flexible and have nice properties for sequential Bayes estimation.
The KL minimization gives the update equations for the parameters of the approximate posterior.

\subsection{Motion Model}
The diffusion process leaves the state mean unchanged and propagates the covariance. 
How much process noise is added to the evolving state is determined by the trace of the covariance matrix (sum of the eigenvalues): each incoming event adds white noise to the covariance diagonal, thus increasing its trace, up to some allowed maximum. 
This works gracefully across many motion speeds. More specifically, we used a maximum standard deviation of 0.03 for poses parametrized in normalized coordinates (with translation in units relative to the mean scene depth), to factor out the metric scale in the diffusion process.

\subsection{Measurement Model}
\label{sec:MeasurementModel}
Here we elaborate on the choice of likelihood function $p(\obs_{k}|\state_{k})$ in~\eqref{eq:BayesCorrectionStepMargPostApprox} that is used to model the events.
Our contributions are, starting from an ideal sensor model, $i$) to define a dimensionless implicit function based on the contrast residual to measure how well the event camera pose and the a priori information (\eg a map of the scene) explain an event (Section~\ref{sec:IdealSensorModel}), 
and $ii$) to build upon such measurement function taking into account noise and outliers, yielding a mixture model for the likelihood function (Section~\ref{sec:MixtureLikelihoodFunction}).

\subsubsection{Ideal Sensor Model}
\label{sec:IdealSensorModel}
In a noise-free scenario, an event is triggered as soon as the temporal contrast reaches the threshold~\eqref{eq:EventTrigCondition}. 
Such a measurement would satisfy $\Delta\ln I-\thr=0$.
For simplicity, let us assume that the polarity has already been taken into account to select the appropriate threshold $\thr^+>0$ or $\thr^-<0$.
Defining the measurement function by
\begin{equation}
\label{eq:Measurementfunction}
M\coloneqq \frac{\Delta\ln I}{\thr} - 1, 
\end{equation}
the event-generation condition becomes $M=0$ in a dimensionless formulation.
Assuming a prediction of the temporal contrast is generated using the system state, $\Delta \ln I (\state_k)$,
then~\eqref{eq:Measurementfunction} depends on both the system state and the observation, $M(\obs_k,\state_k)$.
More precisely, denoting by 
\begin{equation}
\label{eq:stateVectorForContrastCalc}
\tilde{\state} = (\bxi_c,\bxi_{i},\bxi_{j},\thr)^\top,
\end{equation} 
the part of the state~\eqref{eq:StateVecGeneric} needed to compute~\eqref{eq:Measurementfunction}, 
we have $M(\obs_k,\tilde{\state}_k)$.
The likelihood function 
that characterizes such an ideal sensor model is 
\begin{equation}
\label{eq:LikelihoodDelta}
p(\obs_k|\state_k)=\delta(M(\obs_k,\tilde{\state}_k)),
\end{equation}
where $\delta$ is the Dirac delta distribution.

All deviations from ideal conditions can be collectively modeled by a noise term in the likelihood function.
Hence, a more realistic yet simple choice than~\eqref{eq:LikelihoodDelta} that is also supported by the bell-shaped form of the threshold variations observed in the DVS~\cite{Lichtsteiner08ssc} is a Gaussian distribution,
\begin{equation}
\label{eq:LikelihoodGaussian}
p(\obs_k|\state_k) = \cN(M(\obs_k,\tilde{\state}_k);0,\varThr).
\end{equation}

Most previous works in the literature do not consider an implicit measurement function~\eqref{eq:Measurementfunction} or Gaussian model~\eqref{eq:LikelihoodGaussian}
based on the contrast residual.
Instead, they use explicit measurement functions that evaluate the goodness of fit of the event either in the spatial domain (reprojection error)~\cite{Weikersdorfer12robio,Mueggler14iros} or in the temporal domain (event-rate error), \eg image reconstruction thread of~\cite{Kim14bmvc}, assuming Gaussian errors. 
Our measurement function~\eqref{eq:Measurementfunction} is based on the event-generation process and combines in a scalar quantity all the information contained in an event (space-time and polarity) 
to provide a measure of its fit to a given state and a priori information.
However, models based on a single Gaussian distribution~\eqref{eq:LikelihoodGaussian} are very susceptible to outliers.
Therefore, we opt for a mixture model to explicitly account for them, as explained next.

\subsubsection{Resilient Sensor Model. Likelihood Function}
\label{sec:MixtureLikelihoodFunction}
Based on the empirical observation that there is a significant amount of outliers in the event stream, 
we propose a likelihood function consisting of a normal-uniform mixture model.
This model is typical of robust sensor fusion problems~\cite{Vogiatzis11jivc}, where the output of the sensor is modeled as a distribution that mixes a good measurement (normal) with a bad one (uniform):
\begin{align}
\label{eq:likelihood2a}
p(\obs_k|\state_k) = &\, \inlierProb\,\cN(M(\obs_k,\tilde{\state}_k);0,\varThr)\\
\nonumber  & + (1-\inlierProb)\,\cU(M(\obs_k,\tilde{\state}_k); M_{\min},M_{\max}),
\end{align}
where $\inlierProb$ is the inlier probability (and $(1-\inlierProb)$ is the outlier probability).
Inliers are normally distributed around 0 
with variance $\varThr$.
Outliers are uniformly distributed over a known interval $[M_{\min},M_{\max}]$. 
The measurement parameters $\varThr$ and $\inlierProb$ are 
considered unknown and are collected in the state vector $\state_{k}$ to be estimated.

To evaluate $M(\obs_k,\tilde{\state}_k)$, we need to compute the contrast $\Delta \ln I(\tilde{\state}_k)$ in~\eqref{eq:Measurementfunction}. 
We do so based on a known reference image $\Iref$ (and its pose) and both relevant event camera poses for contrast calculation, as explained in Fig.~\ref{fig:TransferEvents}.
Assuming the depth of the scene is known, the point $\bu'$ in the reference image corresponding to the event location $(\bu,t)$ in the event camera satisfies the following equation (in calibrated camera coordinates): 
\begin{equation}
\label{eq:transferredPoint}
\bu'(t) = \pi\bigl( T_{RC}(t) \,\pi^{-1}\bigl(\bu,Z(t)\bigr) \bigr),
\end{equation}
where $T_{RC}(t)$ is the transformation from the event camera frame at time $t$ to the frame of the reference image, $Z(t)$ represents the scene structure (i.e., the depth of the map point corresponding to $\bu$ with respect to the event camera), $\pi:\mathbb{R}^3\to\mathbb{R}^2,\, (X,Y,Z)\mapsto(X/Z,Y/Z)$ is the canonical perspective projection, 
and $\pi^{-1}$ is the inverse perspective projection. 
The transformation $T_{RC}(t_k)$ at the time of the current event depends on the current estimate of the event camera pose $\bxi_c\equiv\bxi(t_k)$ in~\eqref{eq:stateVectorForContrastCalc}; 
the poses $\bxi_i\equiv\bxi(t_i)$ and $\bxi_j\equiv\bxi(t_j)$ along the event camera trajectory $\bxi(t)$ enclosing the past timestamp $t_k-\Delta t$ are used to interpolate the pose $\bxi(t-\Delta t)$, which determines $T_{RC}(t_k-\Delta t)$. 
For simplicity, separate linear interpolations for position and rotation parameters (exponential coordinates) are used, although a Lie Group formulation with the $SE(3)$ exponential and logarithm maps (more computationally expensive) could be used.

Once the corresponding points of the event coordinates $(\bu, t_k)$ and $(\bu, t_k-\Delta t)$ have been computed, we use their intensity values on the reference image $I^r$ to approximate the contrast:
\begin{equation}
\label{eq:ApproxContrastOnRefImage}
\Delta \ln I \approx \ln \Iref(\bu'(t_k)) - \ln \Iref(\bu'(t_k-\Delta t)),
\end{equation}
where $t_k$ is the time of the current event and $\Delta t$ is the time since the last event at the same pixel.
This approach is more accurate than linearizing $\Delta \ln I$. 
We assume that for a small pose change there is a relatively large number of events from different pixels. 
In this case, the information contribution of a new event to an old pose will be negligible, and the new event will mostly contribute to the most recent pose. 

Next, we linearize the measurement function in~\eqref{eq:likelihood2a} around the expected state 
$\bar{\state}_k = E_{p(\state_k|\obs_{1:k-1})}[\state_k]$, 
prior to incorporating the measurement correction:
\begin{align}
\nonumber M(\obs_k,\tilde{\state}_k) & \approx M(\obs_k,\bar{\tilde{\state}}_k) + \nabla_{\tilde{\state}}M(\obs_k,\bar{\tilde{\state}}_k)\cdot(\tilde{\state}_k-\bar{\tilde{\state}}_k) \\
  & = \Mo + \JacMeas \cdot\Delta \tilde{\state}_k,
  \label{eq:M2}
\end{align}
where $\Mo$ and $\JacMeas$ are the predicted measurement and Jacobian at $\bar{\state}_k$, respectively.
Substituting \eqref{eq:M2} in~\eqref{eq:likelihood2a} we get:
\begin{equation}
  p(\obs_k|\state_k) = \inlierProb\,\cN(\Mo+\JacMeas\cdot\Delta \tilde{\state}_k; 0,\varThr)
  +(1-\inlierProb)\,\cU.
  \label{eq:likelihood2b}
\end{equation}
We assume that the linearization is a good approximation to the original measurement function.

Finally, we may re-write the likelihood~\eqref{eq:likelihood2b} in a more general and convenient form for deriving the filter equations, as a sum of exponential families for the state parameters $\state_k$ (see the Appendix
): 
\begin{equation}
  p(\obs_k|\state_k) = \sum_j h(\state_k) \exp(\eta_{o,j}\cdot \suffStat(\state_k) - A_{o,j}).
  \label{eq:family_sum}
\end{equation}

\subsection{Posterior Approximation and Filter Equations}
\label{sec:PosteriorApprox}
Our third contribution pertains to the approximation of the posterior distribution using a tractable distribution.
For this, we consider variational inference theory~\cite{Bishop06book}, and choose a distribution in the exponential family as well as conjugate priors, minimizing the relative entropy error in representing the true posterior distribution with our approximate distribution, as we explain next.

Exponential families of distributions 
are useful in Bayesian estimation because they have \emph{conjugate priors}~\cite{Bishop06book}: if a given distribution is multiplied by a suitable prior, the resulting posterior has the same form as the prior.
Such a prior is called a conjugate prior for the given distribution.
The prior of a distribution in the exponential family is also in the exponential family, which clearly simplifies recursion.
A mixture distribution like~\eqref{eq:family_sum} does not, however, have a conjugate prior:
the product of the likelihood and a prior from the exponential family is not in the family.
Instead, the number of terms of the posterior doubles for each new measurement, making it unmanageable.
Nevertheless, for tractability and flexibility, we choose as conjugate prior a distribution in the exponential family and approximate the product, in the sense of the Kullback-Leibler (KL) divergence~\cite{kullback51aas},
by a distribution of the same form, as expressed by~\eqref{eq:BayesCorrectionStepMargPostApprox}.
This choice of prior is optimal if either the uniform or the normal terms of the likelihood dominates the mixture; we expect that small deviations from this still gives good approximations.

Letting the KL divergence (or relative entropy) from a distribution $f$ to a distribution $g$ be
\begin{equation}
\label{eq:KLdef}
\KLdiv (f\|g) = \int f(x)\log\frac{f(x)}{g(x)}dx,
\end{equation}
which measures the information loss in representing distribution $f$ by means of $g$,
the posterior parameters $\natparam_k$ are calculated by minimization of the KL divergence from the distribution on the right hand side of~\eqref{eq:BayesCorrectionStepMargPostApprox} to the approximating posterior (left hand side of~\eqref{eq:BayesCorrectionStepMargPostApprox}):
\[
\natparam_{k} = \arg\min_\natparam \KLdiv \bigl( C\,p(\obs_{k}|\state_{k}) q(\state_{k};\natparam_{k-1}) \| q(\state_{k};\natparam) \bigr).
\]

It can be shown~\cite[p.505]{Bishop06book} that for $g$ in the exponential family, the necessary optimality condition $\nabla_\natparam \KLdiv (f\| g)=0$ gives the system of equations (in $\natparam$)
\begin{equation}
\label{eq:NecOptCondSystem}
E_{f(\state)}[\suffStat(\state)] = E_{g(\state)}[\suffStat(\state)],
\end{equation}
\ie the expected sufficient statistics must match.
Additionally, the right hand side of~\eqref{eq:NecOptCondSystem} is 
$\nabla A \equiv \nabla_\natparam A = E_{g(\state)}[\suffStat(\state)]$ since $g$ is in the exponential family.
In our case, $g\equiv q(\state_{k};\natparam)$, $f\propto p(\obs_{k}|\state_{k}) q(\state_{k};\natparam_{k-1})$
and~\eqref{eq:NecOptCondSystem} can also be written in terms of the parameters of~\eqref{eq:family_sum} 
[(3)-(6) in the Appendix],
the log-normalizer~$A$ and its gradient:
\begin{align}
\nonumber 0 = & \sum_j \exp\bigl(A(\eta_{o,j}+\eta_{k-1}) - A(\eta_{k-1}) -A_{o,j})\bigr)\\
& \qquad \times \bigl( \nabla A (\eta_{o,j}+\eta_{k-1}) - \nabla A (\natparam)\bigr).
\label{eq:NecOptCondLogNormalizer}
\end{align}
Equation \eqref{eq:NecOptCondLogNormalizer} describes a system of equations that can be solved for~$\natparam$, yielding the update formula for $\natparam_k$ in terms of $\natparam_{k-1}$ and the current event $\obs_k$. 
For a multivariate Gaussian distribution over the event camera poses, 
explicit calculation of all update rules has the simple form of an Extended Kalman Filter (EKF)~\cite{Kalman60jbe,Thrun05book} weighted by the inlier probability of that event:
\begin{align}
  K_k & = P_{k} \JacMeas^{\top} (\JacMeas P_{k}\JacMeas^{\top}+\varThr)^{-1} \label{eq:kalman_gain}\\
  w_k & = \frac{\inlierProb\cN(\Mo; 0,\varThr)}{\inlierProb\cN(\Mo; 0,\varThr)+(1-\inlierProb)\cU} \label{eq:inlier_weight}\\
  \pose_{k+1} & = \pose_{k} + w_k K_k \Mo \label{eq:kalman_update_pose}\\
  P_{k+1} & = (\identity - w_k K_k \JacMeas)P_{k},
  \label{eq:kalman_update_pose_cov}
\end{align}
where $\identity$ is the identity, $\Mo$ and $\JacMeas$ are given in~\eqref{eq:M2}, 
$\pose$ are the \mbox{6-DOF} coordinates (3 for translation and 3 for rotation) of the event camera pose, 
$P$ is the pose covariance matrix, and $w_k K_k$ acts as the Kalman gain.
A pseudocode of the approach is outlined in Algorithm~\ref{alg:Pesudocode}.

The posterior approximation described in this section allows us to fuse the measurements and update the state-vector efficiently, without keeping multiple hypothesis in the style of particle filters, which would quickly become intractable due to the dimension of the state-vector.

\begin{algorithm}[t]
\noindent Initialize state variables (event camera pose, contrast threshold, inlier ratio).
Then, for each incoming event:

\noindent- propagate state covariance (zero-mean random diffusion)

\noindent- transfer the event to the map, compute the depth and evaluate the measurement function $M$ function~\eqref{eq:M2}.

\noindent- compute $K_k$ in~\eqref{eq:kalman_gain}, the inlier probability $\pi_m$, the weight $w_k$ in~\eqref{eq:inlier_weight}, and the gain $w_k K_k$.

\noindent- update filter variables and covariance (e.g., \eqref{eq:kalman_update_pose}-\eqref{eq:kalman_update_pose_cov}).
\caption{\label{alg:Pesudocode} Event-based pose tracking}
\end{algorithm}

\section{Experimental Results}
\label{sec:Evaluation}
Our event-based pose estimation algorithm requires an existing photometric depth map of the scene. 
As mentioned at the beginning of Section~\ref{sec:Methodology}, without loss of generality we describe the map in terms of depth maps with associated reference frames.
These can be obtained from a previous mapping stage by means of an RGB-D camera or by classical dense reconstruction approaches using standard cameras (e.g., DTAM~\cite{Newcombe11iccv} or REMODE~\cite{Pizzoli14icra}), RGB-D sensors~\cite{Newcombe11ismar}, 
or even using an event camera (future research).
In this work we use an Intel Realsense R200 RGB-D camera. 
We show experiments with both nearly planar scenes and scenes with large depth variations.

We evaluated the performance of our algorithm on several indoor and outdoor sequences.
The datasets also contain fast motion with excitations in all six degrees of freedom (DOF). For the interested reader, we would like to point out that sequences similar to the ones used in these experiments can be found in the publicly available Event Camera Dataset~\cite{Mueggler17ijrr}.

\subsubsection*{Indoor Experiments}
\begin{figure*}[!ht]
\centering
\raisebox{-0.4\height}{\includegraphics[width=0.27\linewidth]{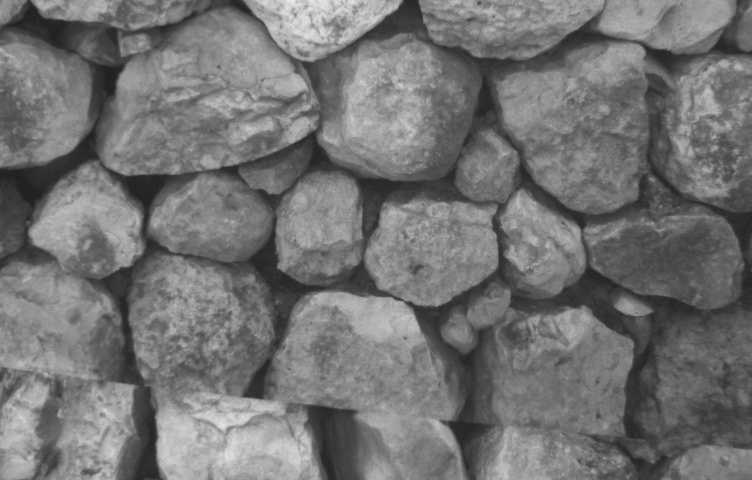}}\;
\raisebox{-0.5\height}{\includegraphics[width=0.35\linewidth]{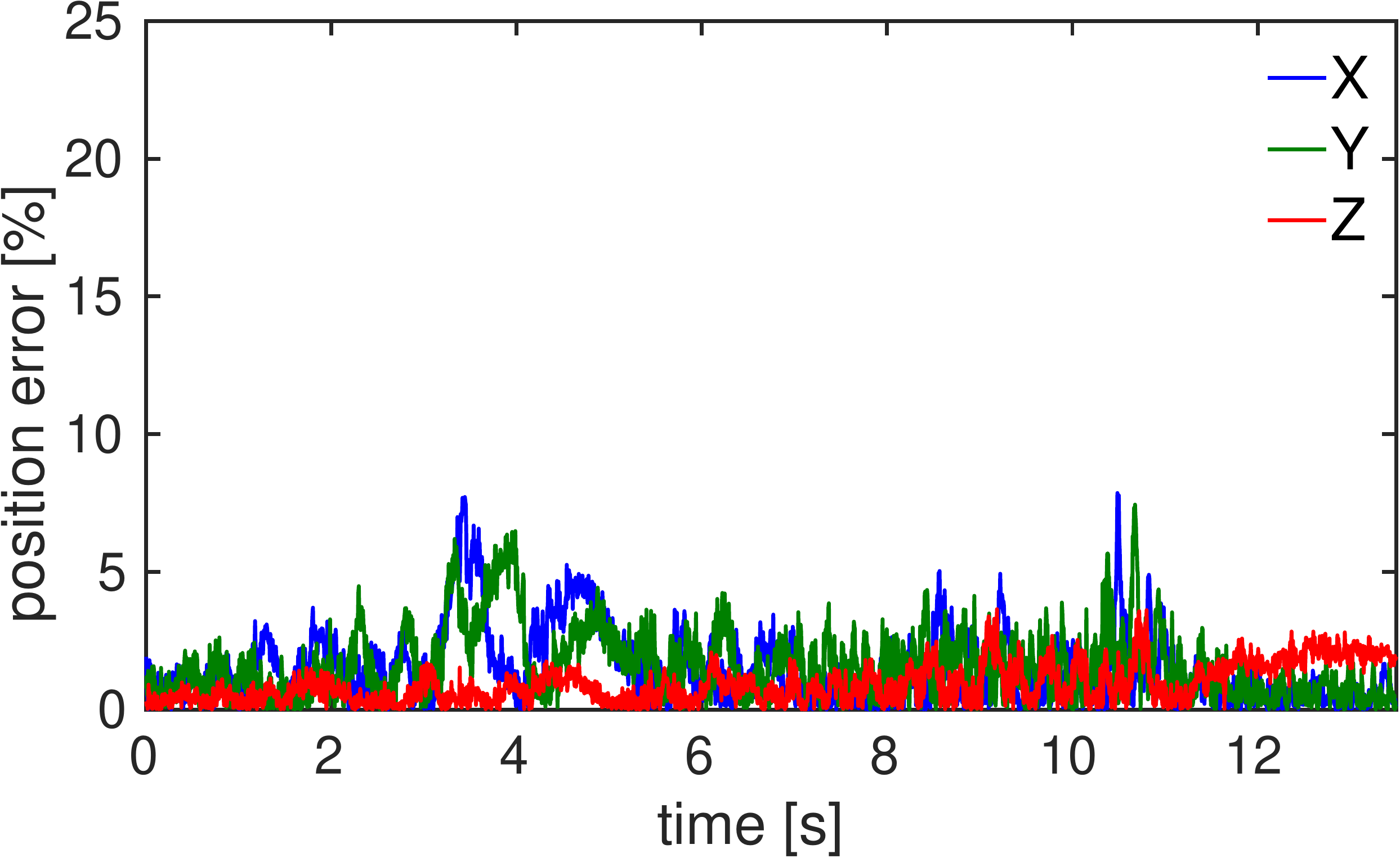}}\;
\raisebox{-0.5\height}{\includegraphics[width=0.35\linewidth]{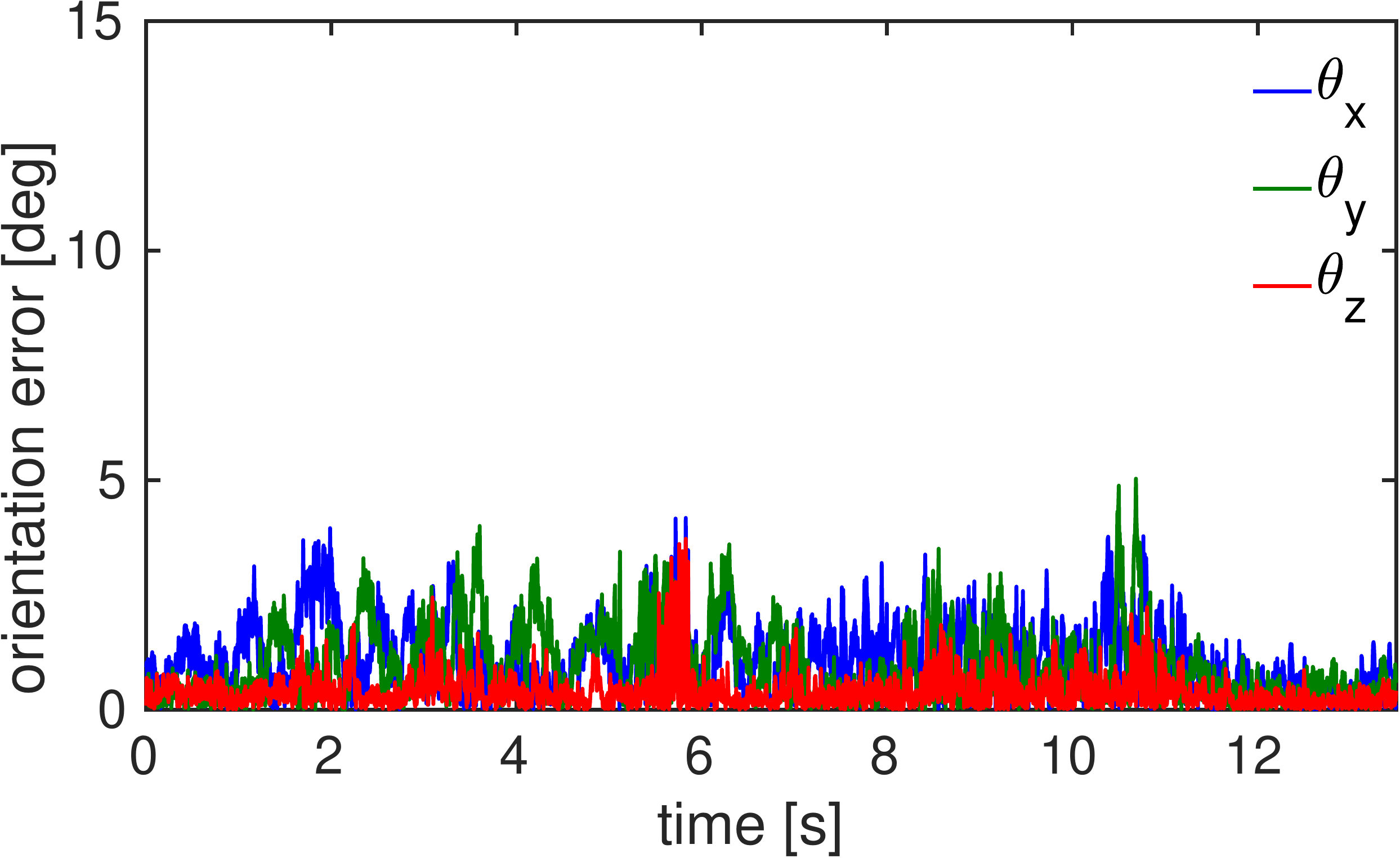}}
   \caption{Error plots in position (relative to a mean scene depth of \SI{60}{\centi\meter}) and in orientation (in degrees) for one of the test sequences with ground truth provided by a motion capture system with sub-millimeter accuracy.
   }
\label{fig:ErrorPlotsOptitrack}
\end{figure*}

\begin{figure}[t!]
\centering
\raisebox{-0.5\height}{\includegraphics[width=0.48\columnwidth]{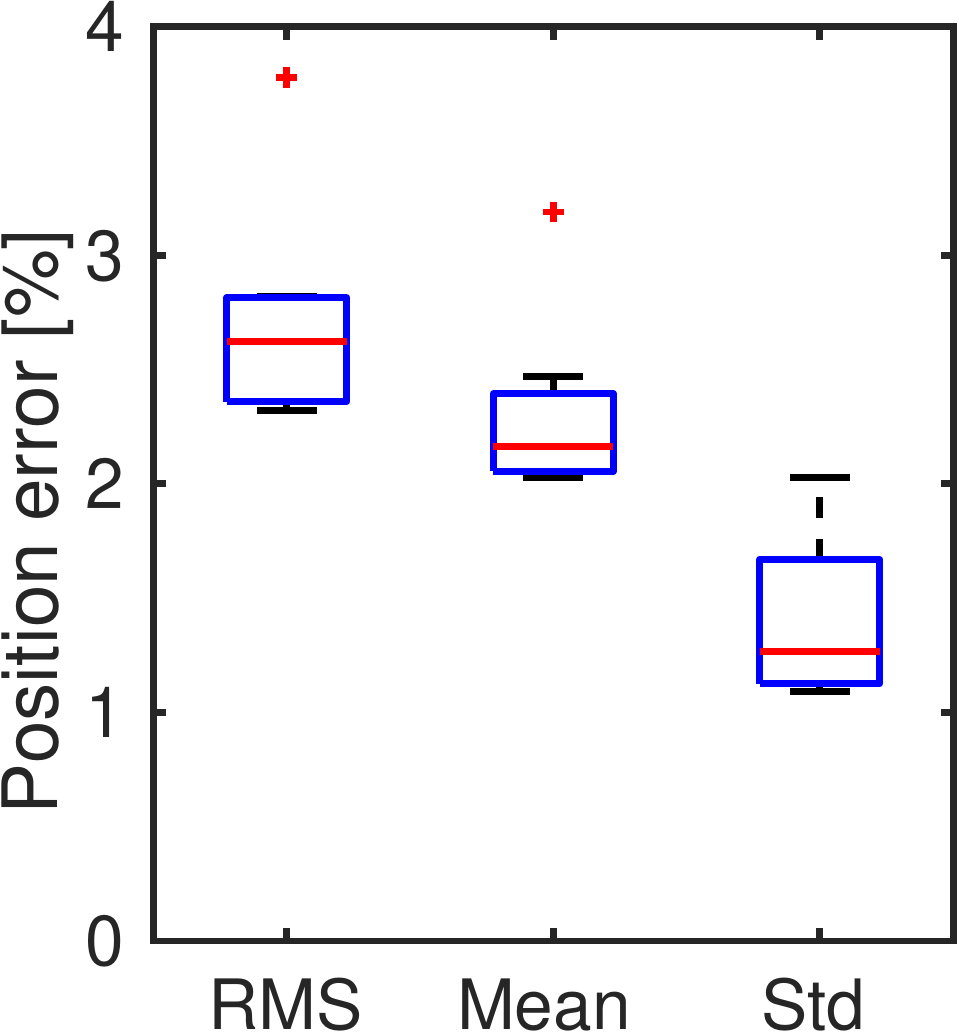}}\;\;\;\raisebox{-0.5\height}{\includegraphics[width=0.48\columnwidth]{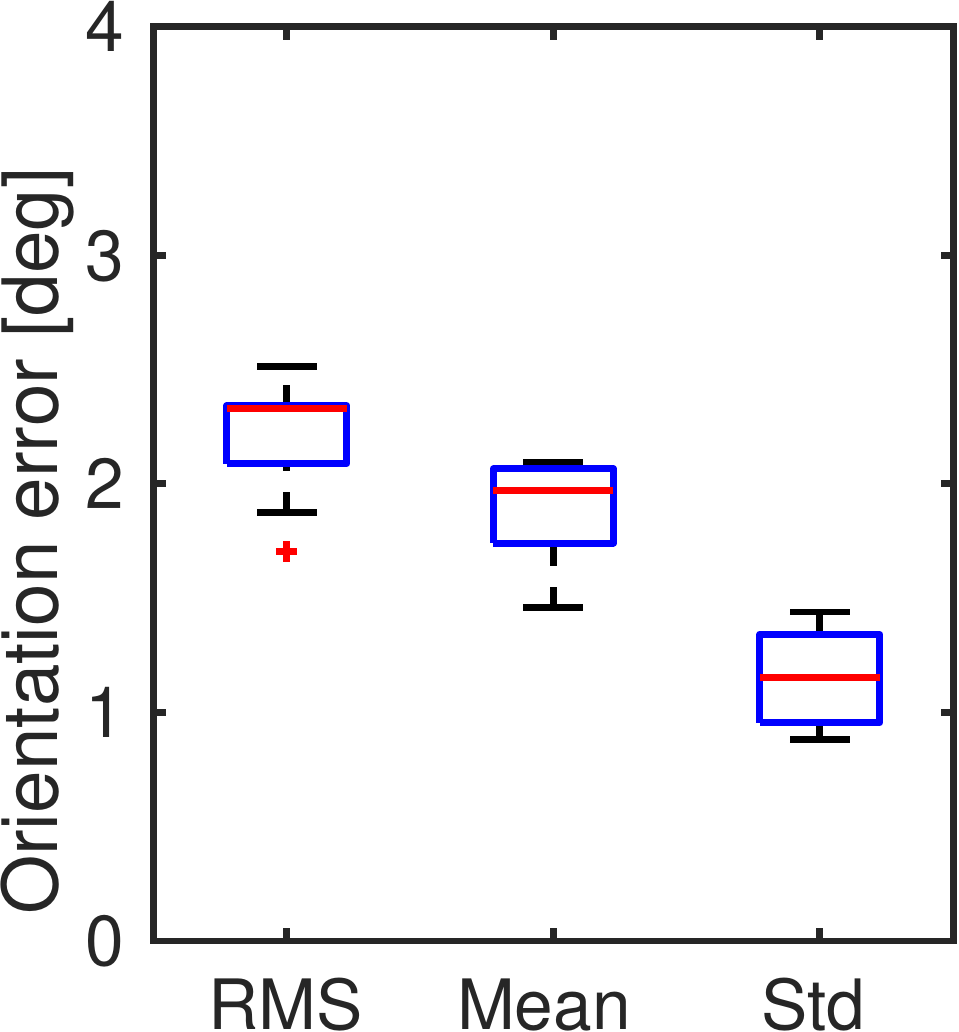}}
   \caption{Error in position (relative to a mean scene depth of \SI{60}{\centi\meter}) and orientation (in degrees) of the trajectories recovered by our method for \emph{all} {\footnotesize\texttt{rocks}} sequences (ground truth is given by a motion capture system).
   We provide box plots of the root-mean-square (RMS) errors, the mean errors and the standard deviation (Std) of the errors.}
\label{fig:BoxPlotsOptitrack}
\end{figure}

First, we assessed the accuracy of our method against ground truth obtained by a motion-capture system.
We placed the event camera in front of a scene consisting of rocks (Fig.~\ref{fig:ErrorPlotsOptitrack}) at a mean scene depth of \SI{60}{\centi\meter} and recorded eight sequences.
Fig.~\ref{fig:ErrorPlotsOptitrack} shows the position and orientation errors (i.e., difference between the estimated ones and ground truth)\footnote{The rotation error is measured using the angle of their relative rotation (\ie geodesic distance in $SO(3)$~\cite{Huynh09jmiv}).}
for one of the sequences, while Fig.~\ref{fig:indoorsOrientation} shows the actual values of the estimated trajectory and ground truth over time.
Fig.~\ref{fig:BoxPlotsOptitrack} summarizes the errors of the estimated trajectories for all sequences.
The mean RMS errors in position and orientation are \SI{1.63}{\centi\meter} and \SI{2.21}{\degree}, respectively,
while the mean and standard deviations of the position and orientation errors are $\mu = \SI{1.38}{\centi\meter}$, $\sigma = \SI{0.84}{\centi\meter}$, and 
$\mu = \SI{1.89}{\degree}$, $\sigma = \SI{1.15}{\degree}$, respectively.
Notice that the RMS position error corresponds to \SI{2.71}{\percent} of the average scene depth, which is very good despite the poor spatial resolution of the DVS.

\subsubsection*{Outdoor Experiments}
\begin{figure}[t!]
\centering
   \includegraphics[width=0.85\linewidth]{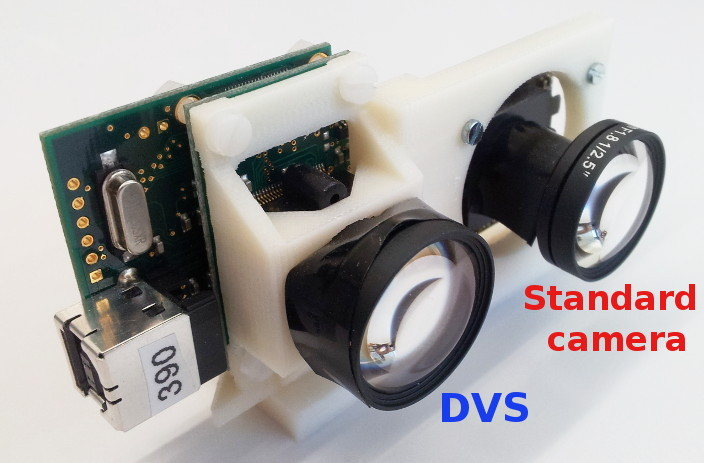}
   \caption{An event camera (DVS) and a standard camera mounted on a rig. The standard camera was only used for comparison.
   }
\label{fig:DVSCameraGun}
\end{figure}

\begin{figure*}[t]
\centering
   \raisebox{-0.4\height}{\includegraphics[width=0.27\linewidth]{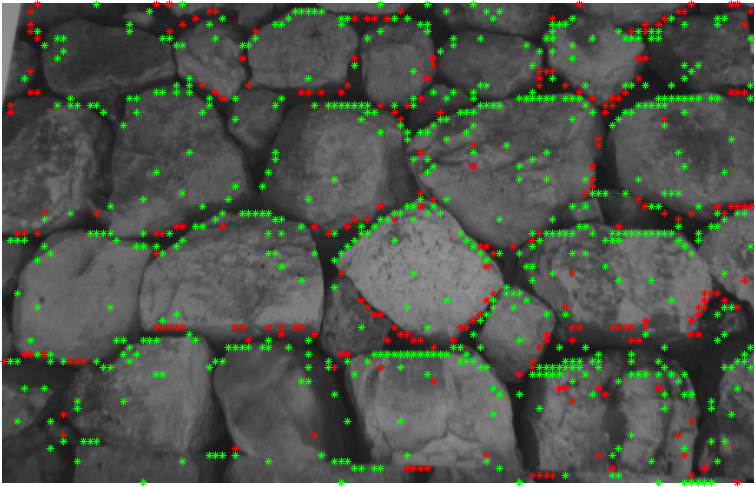}}\;
   \raisebox{-0.5\height}{\includegraphics[width=0.35\linewidth]{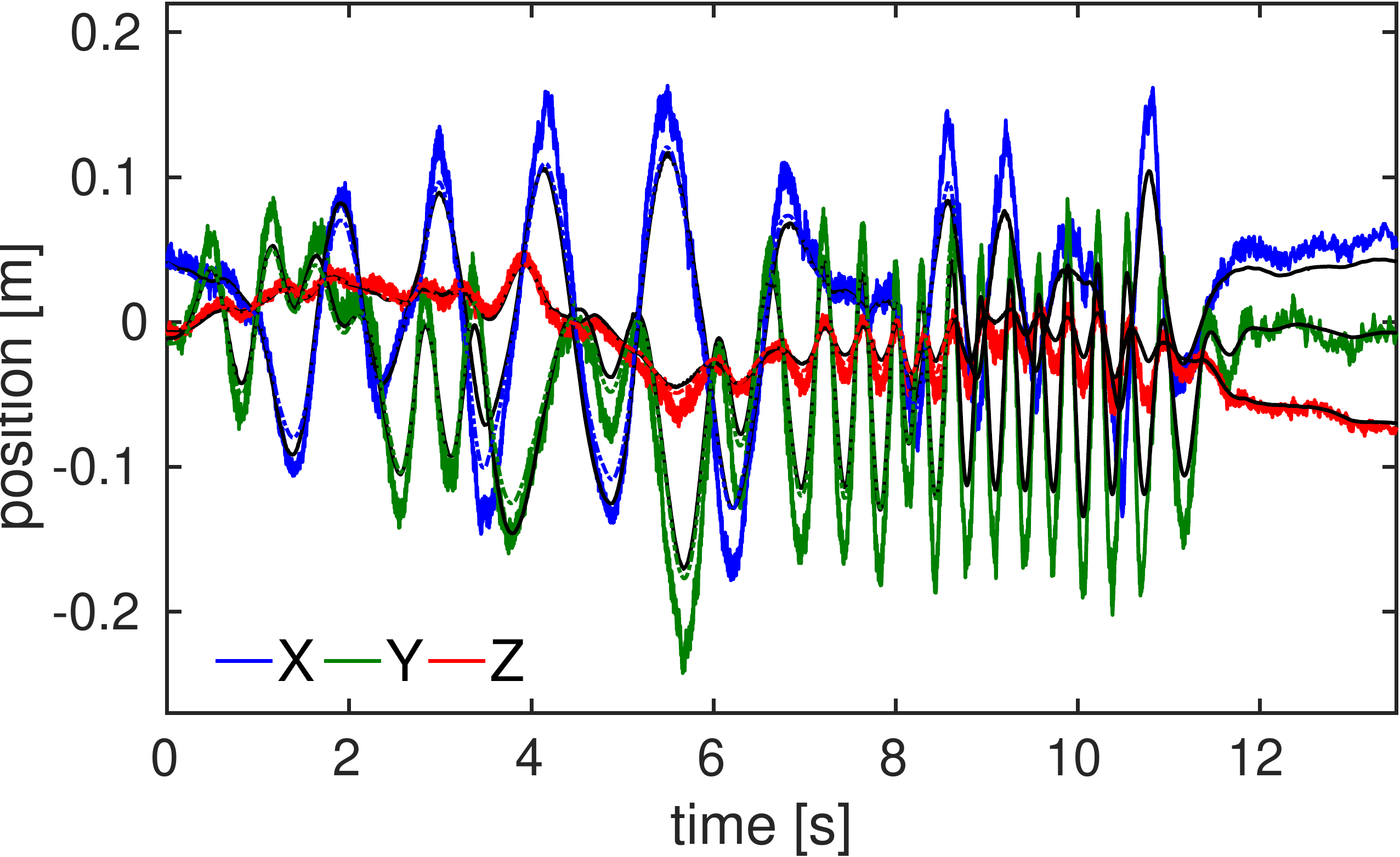}}\;
   \raisebox{-0.5\height}{\includegraphics[width=0.35\linewidth]{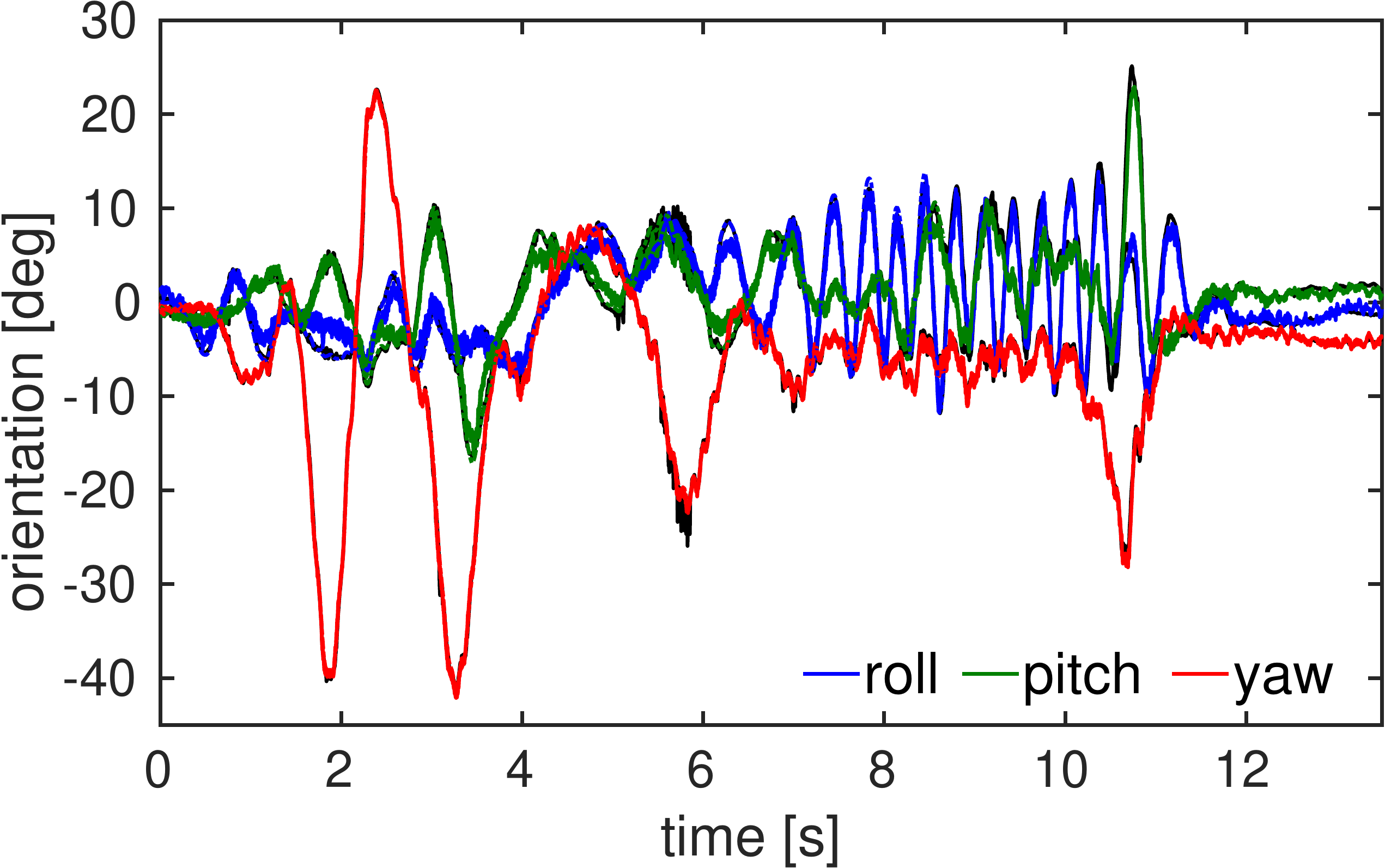}}
   \caption{Indoor experiment with \mbox{6-DOF} motion.
   Left: Image of the standard camera overlaid with events (during mild motion).
   Events are displayed in red and green, according to polarity.
   Estimated position (center) and orientation (right) from our event-based algorithm (solid line), a frame-based method (dash-dot line) and ground truth (black line) from a motion capture system.
   }
\label{fig:indoorsOrientation}
\end{figure*}

For the outdoor experiments, we used structure from motion from a standard camera as ground truth,  
more specifically we used SVO~\cite{Forster14icra}.
To this end, we rigidly mounted the DVS and a standard camera on a rig (see \figurename~\ref{fig:DVSCameraGun}),
and the same lens model was mounted on both sensors. 
The DVS has a spatial resolution of $128\times 128$ pixels and operates asynchronously, in the microsecond scale.
The standard camera is a global shutter MatrixVision Bluefox camera with a resolution of $752\times 480$ pixels and a frame rate of up to \SI{90}{\hertz}.
Both camera and DVS were calibrated intrinsically and extrinsically.
For reference, we measured the accuracy of the frame-based method against the motion-capture system, 
in the same sequences previously mentioned ({\footnotesize\texttt{rocks}}, as in Fig.~\ref{fig:indoorsOrientation}).
The average RMS errors in position and orientation are \SI{1.08}{\centi\meter} (i.e., \SI{1.8}{\percent} of the mean scene depth) and \SI{1.04}{\degree}, respectively.
Comparing these values to those of the event-based method, 
we note that, in spite of the limited resolution of the DVS, 
the accuracy of the results provided by our event-based algorithm is only slightly worse 
(\SI{2.71}{\percent} vs. \SI{1.8}{\percent} in position, and \SI{2.21}{\degree} vs. \SI{1.04}{\degree} in orientation)
than that obtained by a standard camera processing $20\times$ higher resolution images. 
This is made possible by the DVS temporal resolution being ten thousand times larger than the standard camera.

\begin{figure*}[t!]
\centering
\raisebox{-0.4\height}{\includegraphics[width=0.26\linewidth]{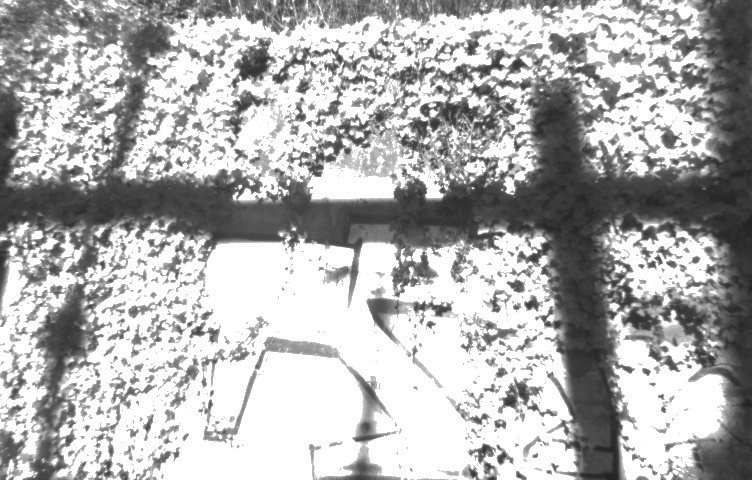}}\;\;\;
\raisebox{-0.5\height}{\includegraphics[width=0.30\linewidth]{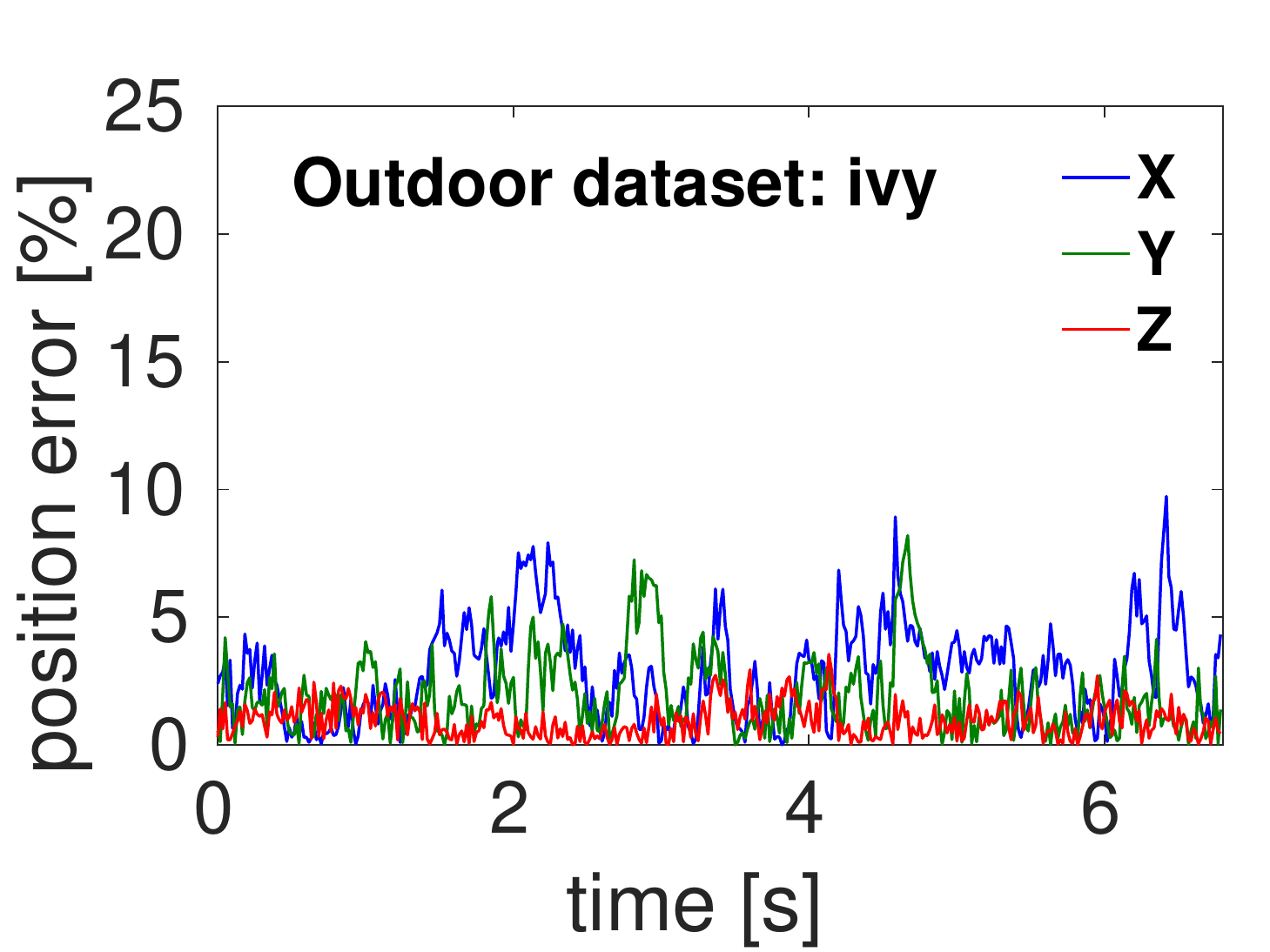}}\;\;\;
\raisebox{-0.5\height}{\includegraphics[width=0.30\linewidth]{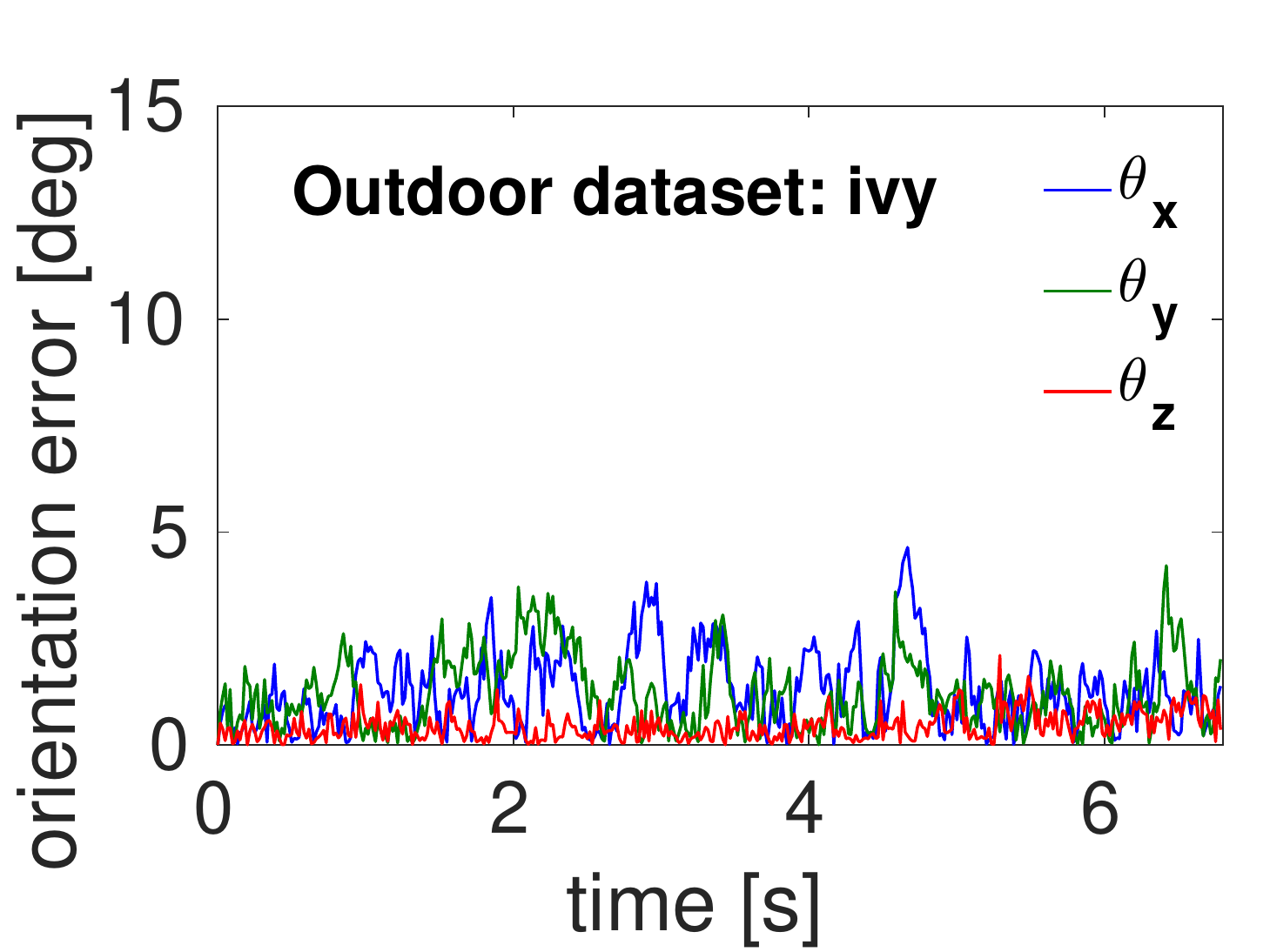}}\\
\raisebox{-0.4\height}{\includegraphics[width=0.26\linewidth]{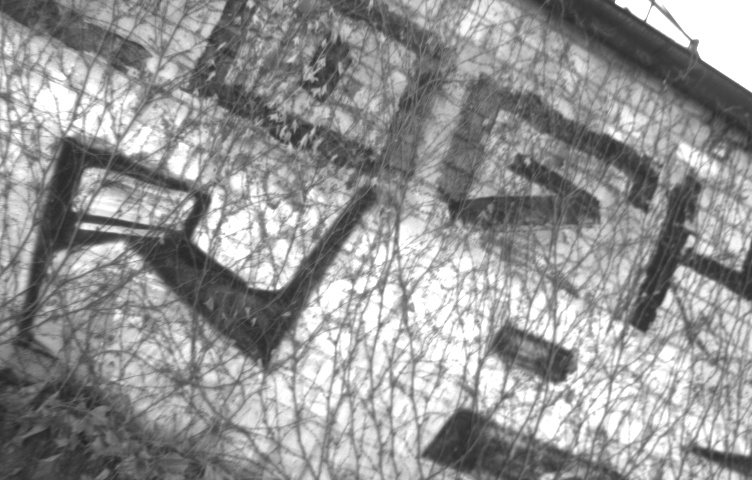}}\;\;\;
\raisebox{-0.5\height}{\includegraphics[width=0.30\linewidth]{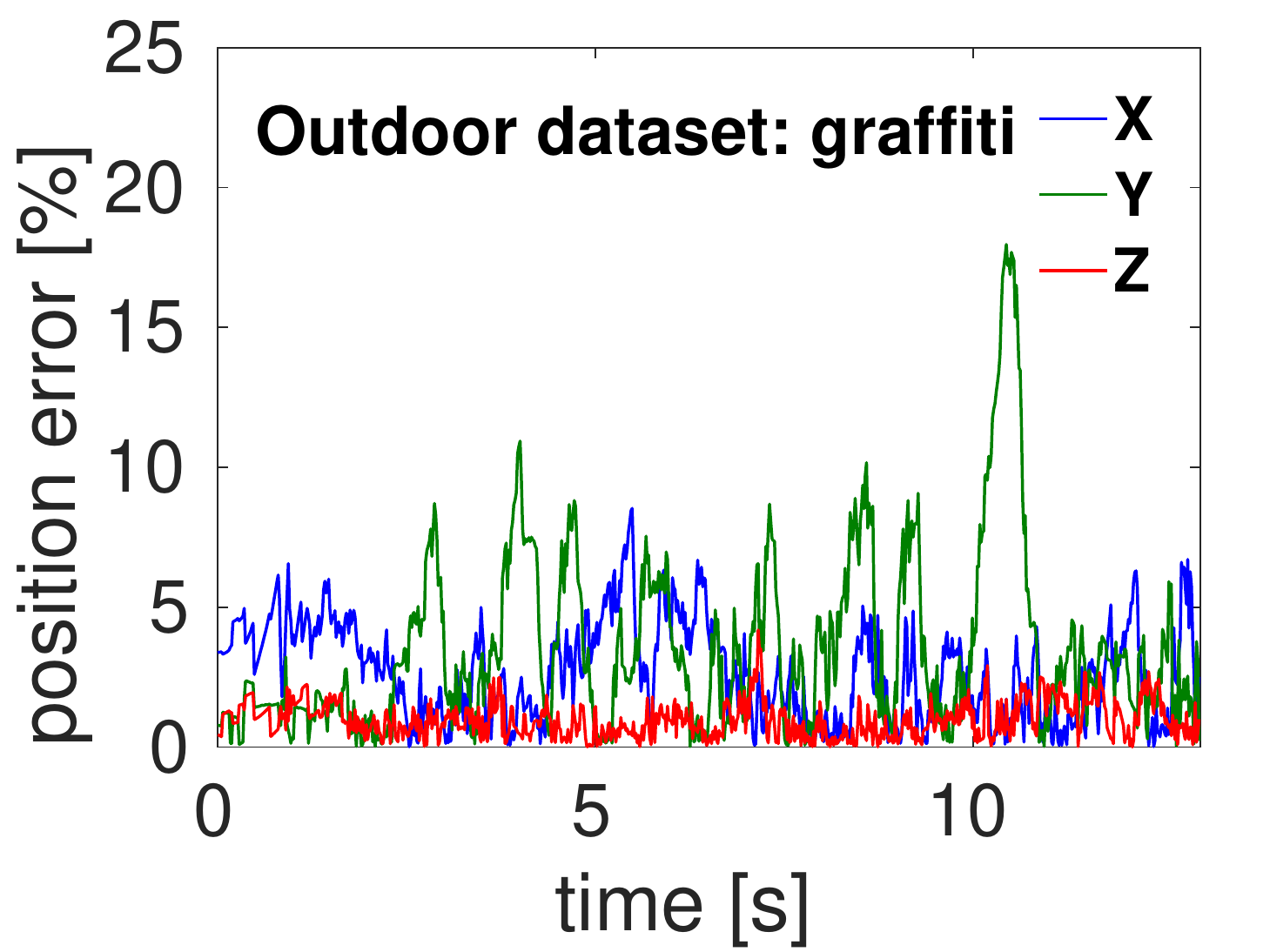}}\;\;\;
\raisebox{-0.5\height}{\includegraphics[width=0.30\linewidth]{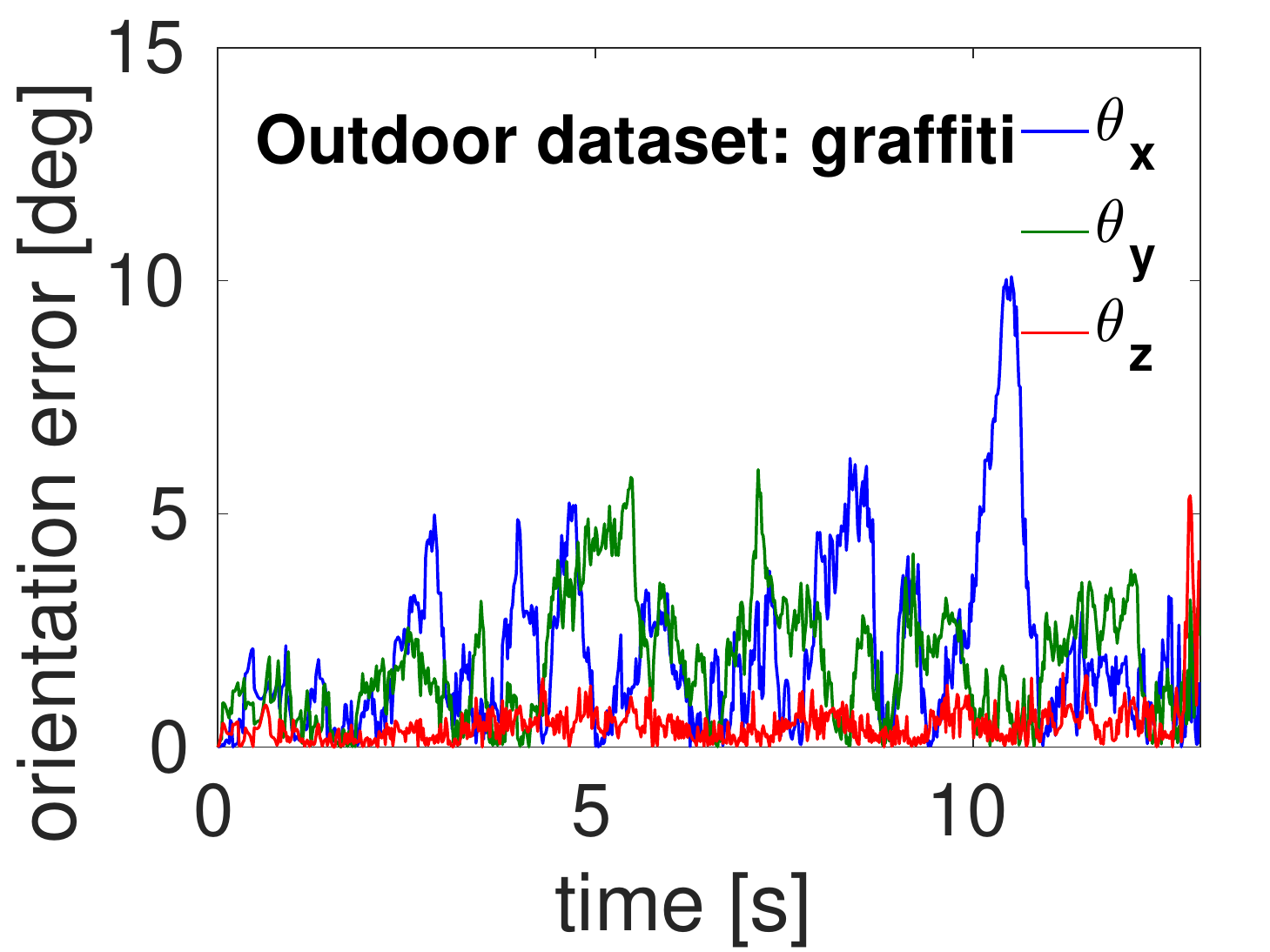}}\\
\raisebox{-0.4\height}{\includegraphics[width=0.26\linewidth]{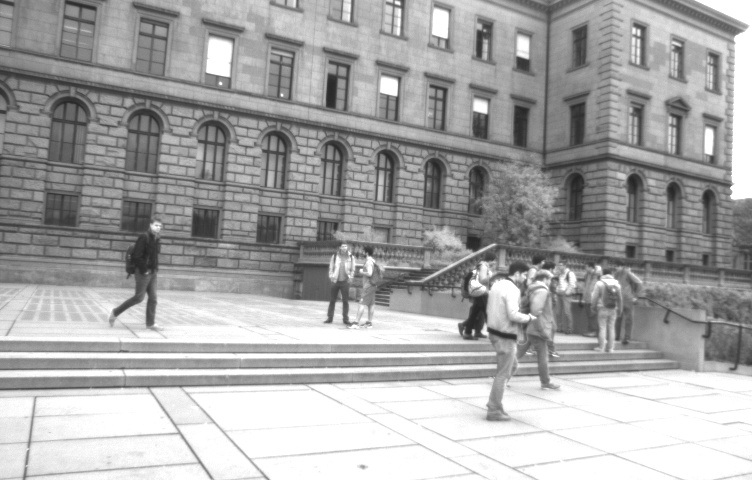}}\;\;\;
\raisebox{-0.5\height}{\includegraphics[width=0.30\linewidth]{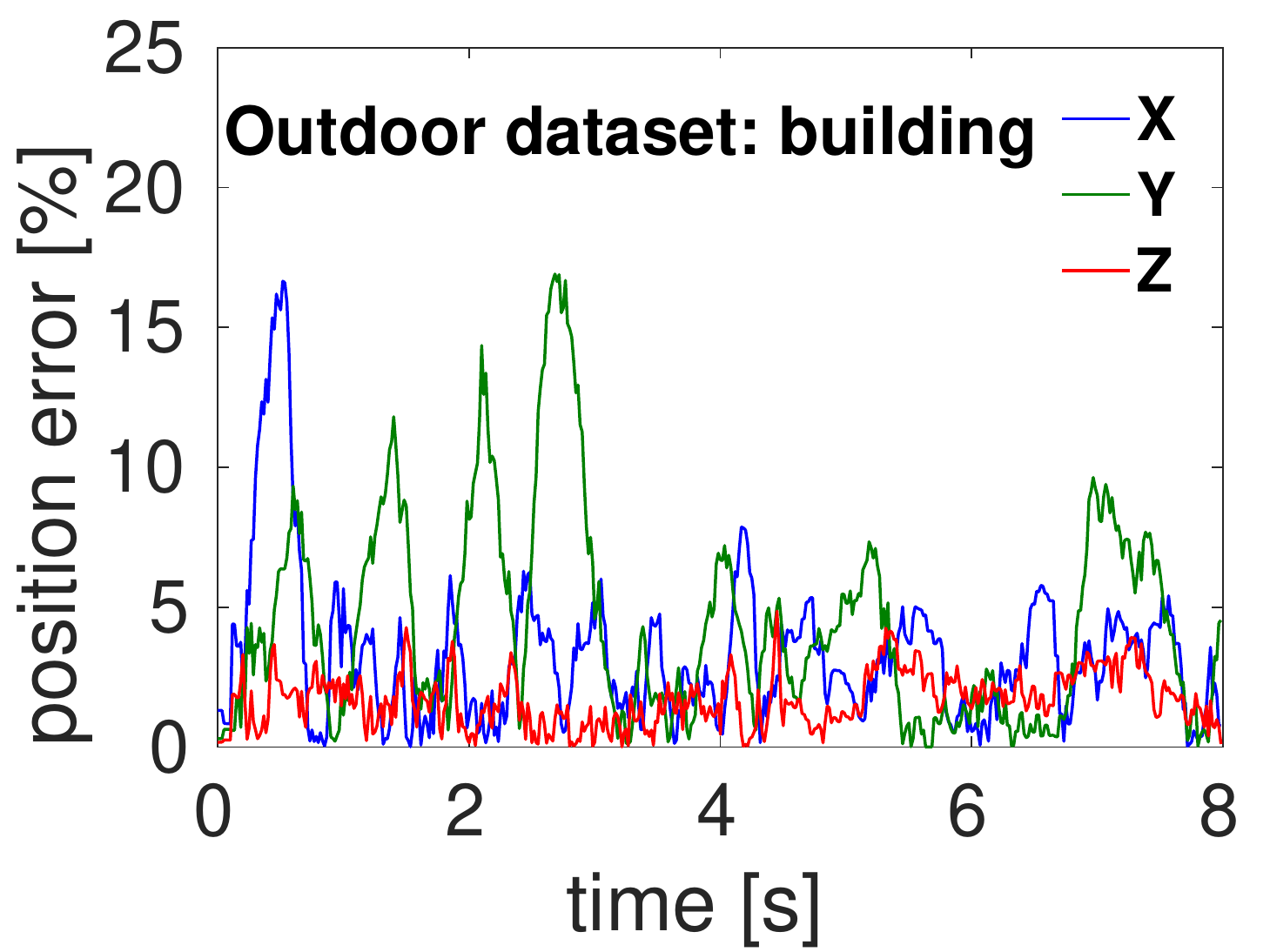}}\;\;\;
\raisebox{-0.5\height}{\includegraphics[width=0.30\linewidth]{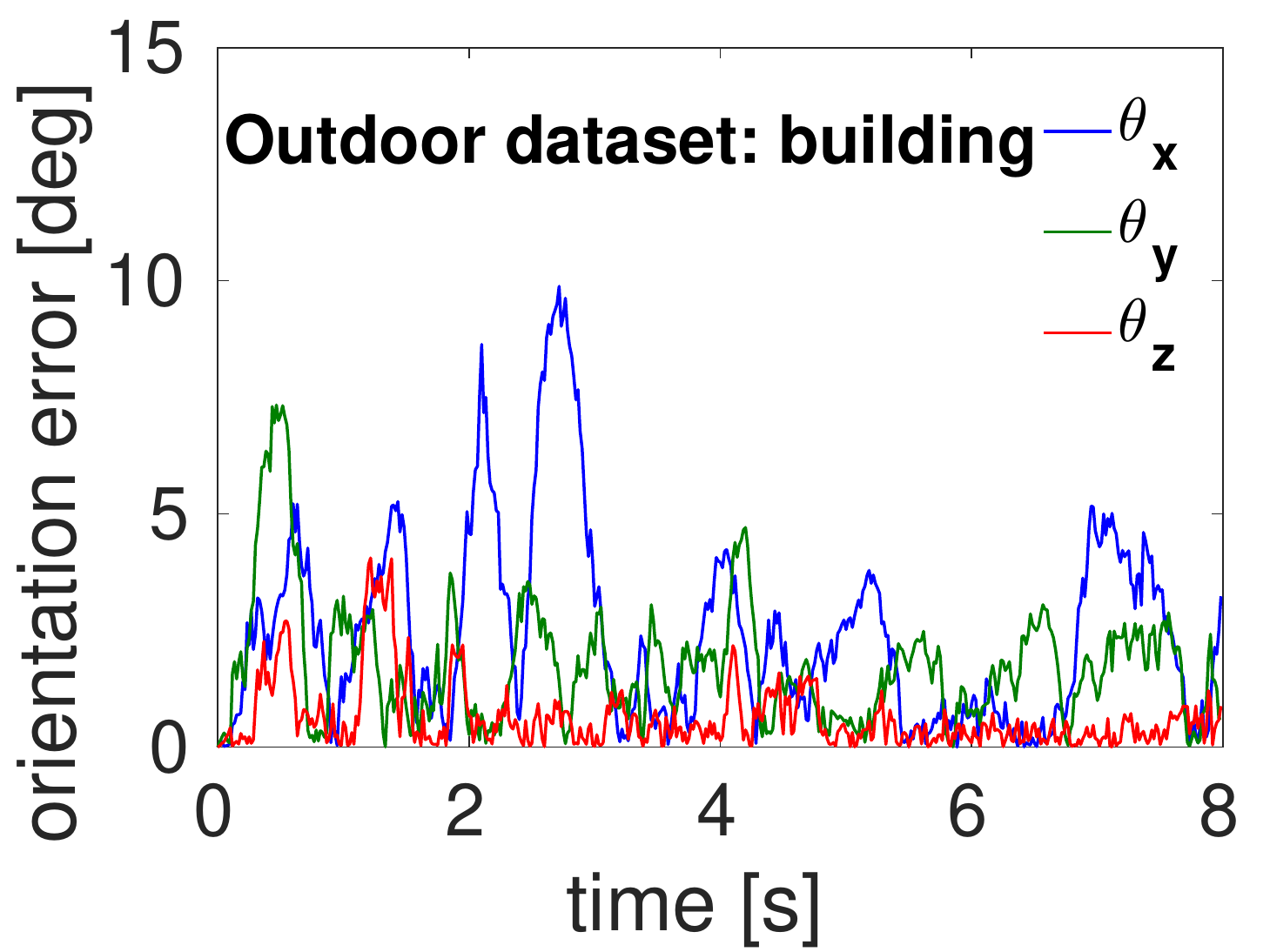}}
   \caption{Error plots in position (2nd column, relative to the mean scene depth) and in orientation (3rd column, in degrees) for three outdoor test sequences (1st column): {\footnotesize\texttt{ivy}}, {\footnotesize\texttt{graffiti}}, and {\footnotesize\texttt{building}}.
   The mean scene depths are \SI{2.5}{\meter}, \SI{3}{\meter}, and \SI{30}{\meter}, respectively.}
\label{fig:ErrorPlots}
\end{figure*}

The three outdoor sequences ({\footnotesize\texttt{ivy}}, {\footnotesize\texttt{graffiti}}, and {\footnotesize\texttt{building}}) 
were recorded with the DVS-plus-camera rig viewing an ivy, a graffiti covered by some plants, and a building with people moving in front of it, respectively (see Fig.~\ref{fig:ErrorPlots}, 1st column and accompanying video submission).
The rig was moved by hand with increasing speed. 
All sequences exhibit significant translational and rotational motion.
The error plots in position and orientation of all 6-DOFs are given in Fig.~\ref{fig:ErrorPlots}.
The reported error peaks in the {\footnotesize\texttt{graffiti}} and {\footnotesize\texttt{building}} sequences are due to a decrease of overlap between the event camera frustum and the reference map, thus making pose estimation ambiguous for some motions (e.g., $Y$-translation vs. $X$-rotation).

\begin{table}[b!]
  \centering
  \caption{Error measurements of three outdoor sequences.
  Translation errors are relative (\ie scaled by the mean scene depth).}
  \label{tab:PoseErrors}
  \begin{tabular}{l|SSS|SSS}
    \toprule
    & \multicolumn{3}{c|}{Position error [\si{\percent}]} & 
    \multicolumn{3}{c}{Orientation error [\si{\degree}]} \\
    & RMS & $\mu$ & $\sigma$ & RMS & $\mu$ & $\sigma$ \\
    \midrule
    {\footnotesize\texttt{ivy}} & 4.37 & 3.97 & 1.84 & 2.21 & 2.00 & 0.94 \\
    {\footnotesize\texttt{graffiti}} & 5.88 & 5.23 & 2.70 & 3.58 & 3.09 & 1.80 \\
    {\footnotesize\texttt{building}} & 7.40 & 6.47 & 3.60 & 3.99 & 3.43 & 2.05\\
    \bottomrule
  \end{tabular}
\end{table}

\begin{figure}[t]
\centering
   \includegraphics[height=5cm]{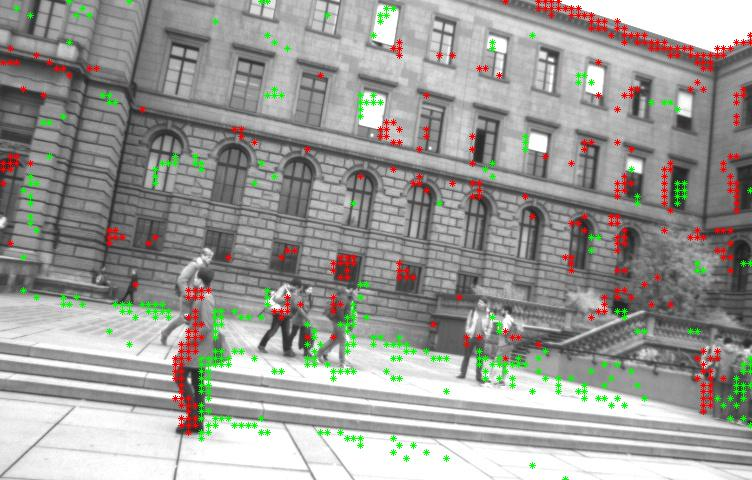}
   \caption{The algorithm is able to track the pose of the event camera in spite of the considerable amount of events generated by moving objects (\eg people) in the scene.\label{fig:outdoorsScreenshot}}
\end{figure}

\begin{figure*}[t]
\centering
   \raisebox{-0.4\height}{\includegraphics[width=0.27\linewidth]{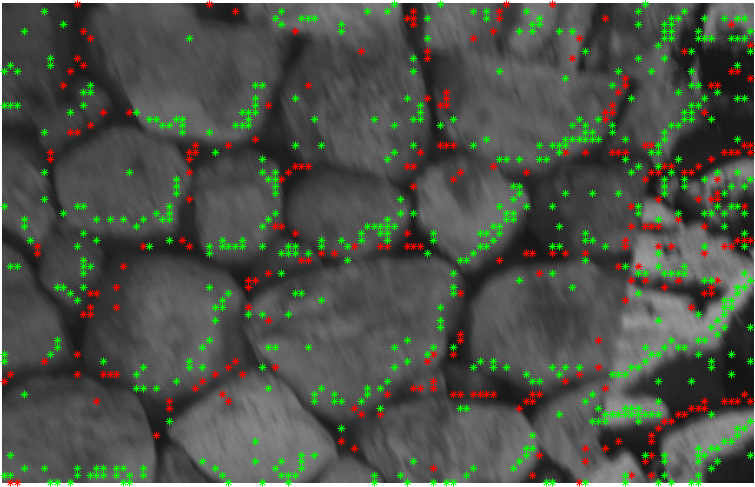}}\;
   \raisebox{-0.5\height}{\includegraphics[width=0.35\linewidth]{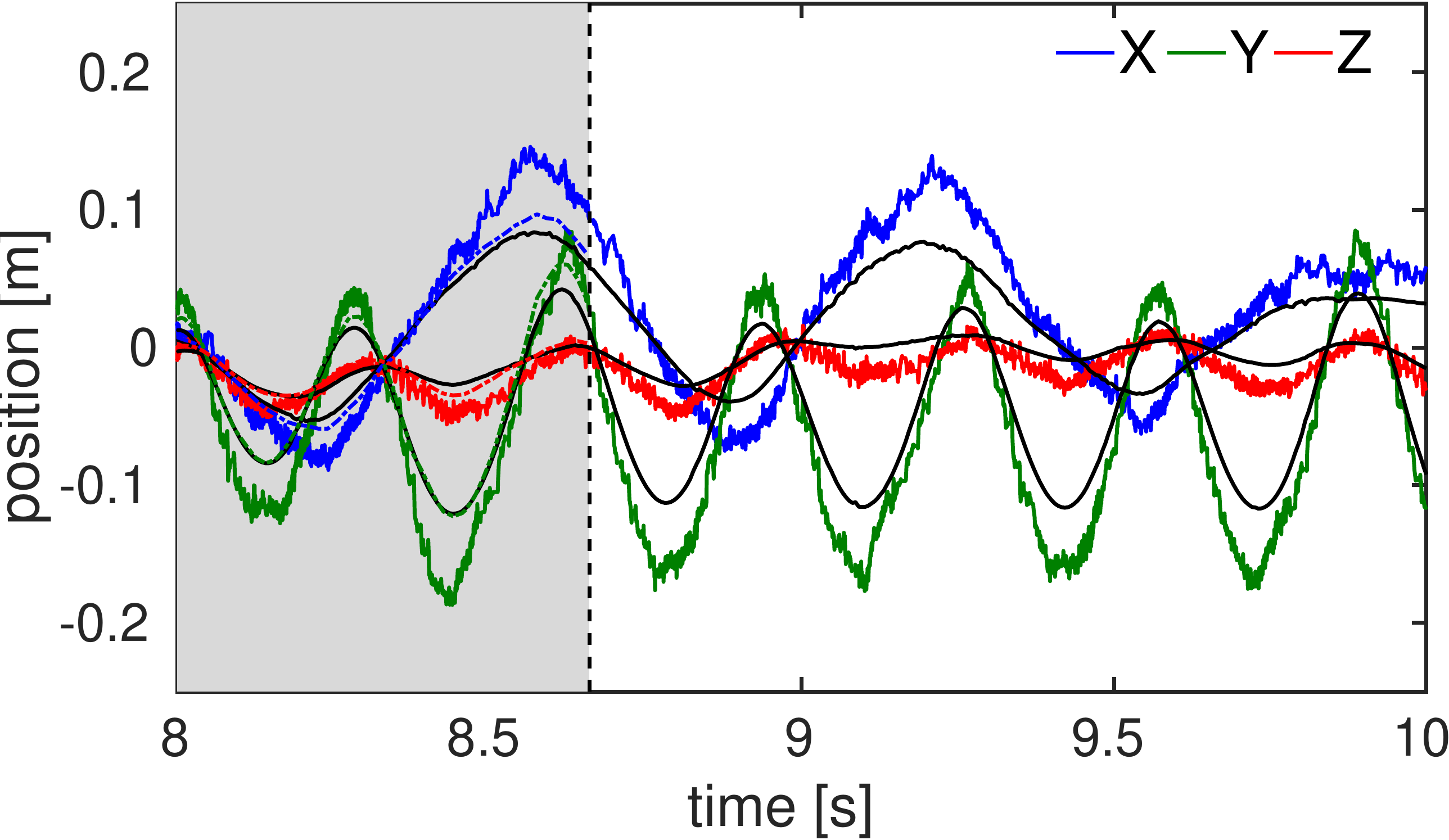}}\;
   \raisebox{-0.5\height}{\includegraphics[width=0.35\linewidth, height=3.7cm]{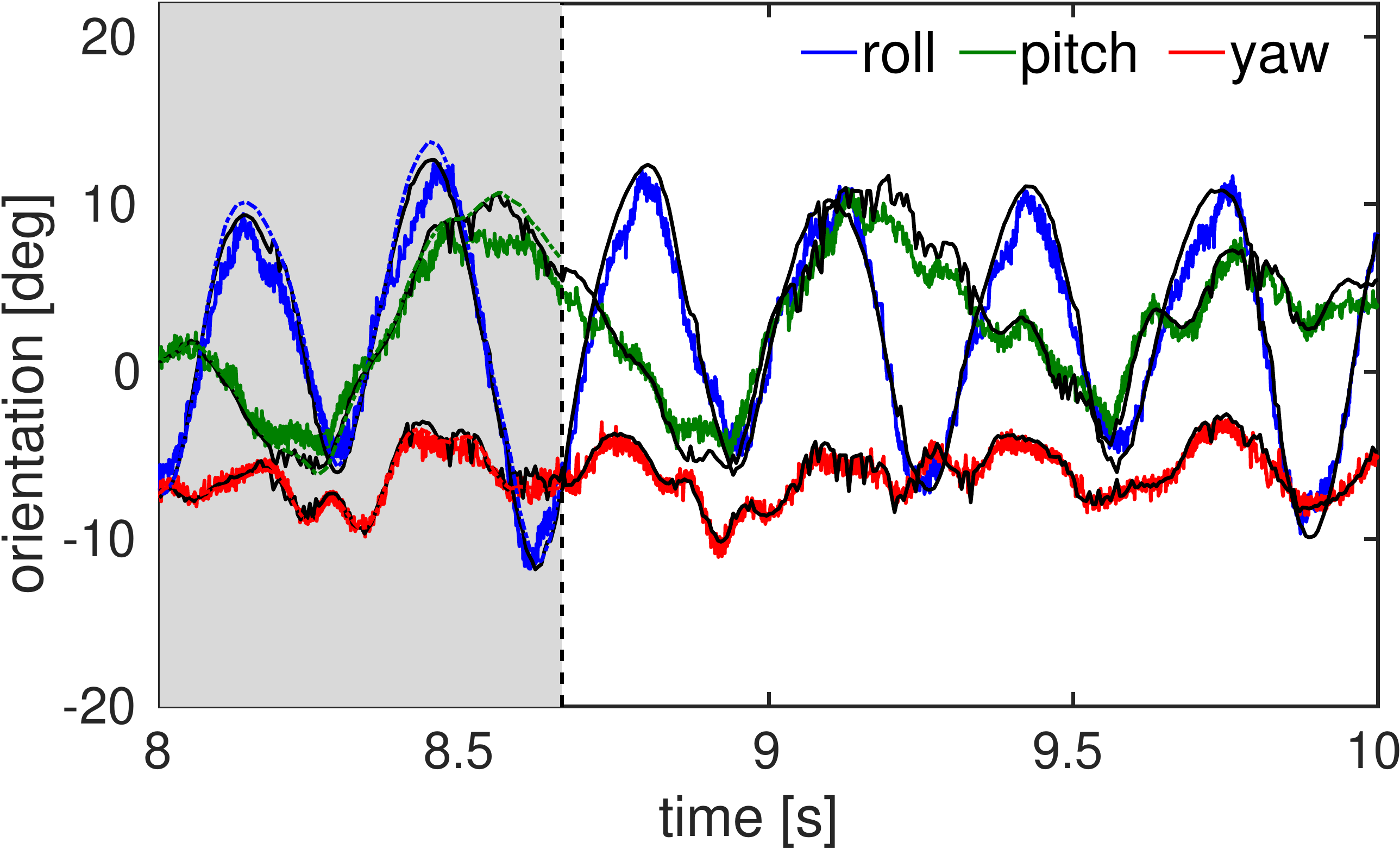}}
   \caption{Zoom of Fig.~\ref{fig:indoorsOrientation}. 
   Left: Image of the standard camera overlaid with events (red and green points, according to polarity) during high-speed motion.
   Center and right: estimated trajectories. 
   Due to the very high temporal resolution, our algorithm can still track the motion even when the images of the standard camera are sufficiently blurred so that the frame-based method (FB) failed.
   The event-based method (EB) provides pose updates even in high-speed motions,
   whereas the frame-based method loses track (it only provides pose updates in the region marked with the shaded area, then it fails).}
\label{fig:indoorsOrientationZoom}
\end{figure*}

Table~\ref{tab:PoseErrors} summarizes the statistics of the pose tracking error for the three outdoor sequences.
For the {\footnotesize\texttt{ivy}} dataset, the mean and standard deviation of the position error are \SI{9.93}{\centi\meter} and \SI{4.60}{\centi\meter}, which correspond to \SI{3.97}{\percent} and \SI{1.84}{\percent} of the average scene depth (\SI{2.5}{\meter}), respectively.
The mean and standard deviation of the orientation error are \SI{2.0}{\degree} and \SI{0.94}{\degree}, respectively.
For the {\footnotesize\texttt{building}} dataset, which presents the largest errors, 
the mean and standard deviation of the orientation error are \SI{3.43}{\degree} and \SI{2.05}{\degree}, respectively,
while, in position error, the corresponding figures are \SI{1.94}{\meter} and \SI{1.08}{\meter}, that correspond to \SI{6.47}{\percent} and \SI{3.60}{\percent} of the average scene depth (\SI{30}{\meter}), respectively.

As reported by the small errors in Table~\ref{tab:PoseErrors}, overall our event-based algorithm is able to accurately track the pose of the event camera also outdoors.
We expect that the results provided by our approach would be even more accurate with the next generation of event-based sensors currently being developed~\cite{Brandli14ssc,Li15iiws}, which will have higher spatial resolution ($640 \times 480$ pixels). 
Finally, observe that in the {\footnotesize\texttt{building}} sequence (Fig.~\ref{fig:ErrorPlots}, bottom row), our method gracefully tracks the pose in spite of the considerable amount of events generated by moving objects (\eg people) in the scene (see~\figurename~\ref{fig:outdoorsScreenshot}).

\subsection{Tracking during High-Speed Motions}

In addition to the error plots in Fig.~\ref{fig:ErrorPlotsOptitrack}, 
we show in Fig.~\ref{fig:indoorsOrientation} the actual values of the trajectories (position and orientation) acquired by the motion capture system (dashed line) and estimated by the event-based method (solid line) and the frame-based method (dash-dot).
Notice that they are all are almost indistinguishable relative to the amplitude of the motion excitation, 
which gives a better appreciation of the small errors reported in Figs.~\ref{fig:ErrorPlotsOptitrack} and~\ref{fig:BoxPlotsOptitrack}.

Figure~\ref{fig:indoorsOrientationZoom} shows a magnified version of the estimated trajectories during high-speed motions (occurring at $t\geq \SI{7}{\second}$ in Fig.~\ref{fig:indoorsOrientation}).
The frame-based method is able to track in the shaded region, up to $t\approx \SI{8.66}{\second}$ (indicated by a vertical dashed line), at which point
it loses tracking due to motion blur, while our event-based method continues to accurately estimate the pose.

\subsection{Experiments with Large Depth Variation}

\begin{figure*}[t!]
\centering
\raisebox{-0.4\height}{\includegraphics[width=0.26\linewidth]{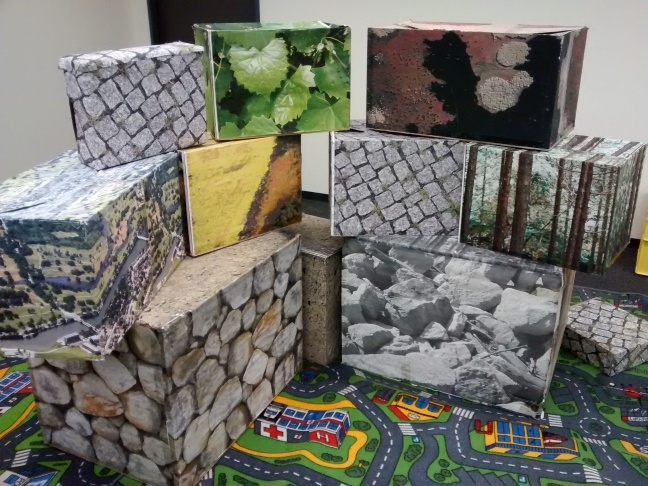}}\;\;\;
\raisebox{-0.5\height}{\includegraphics[width=0.34\linewidth]{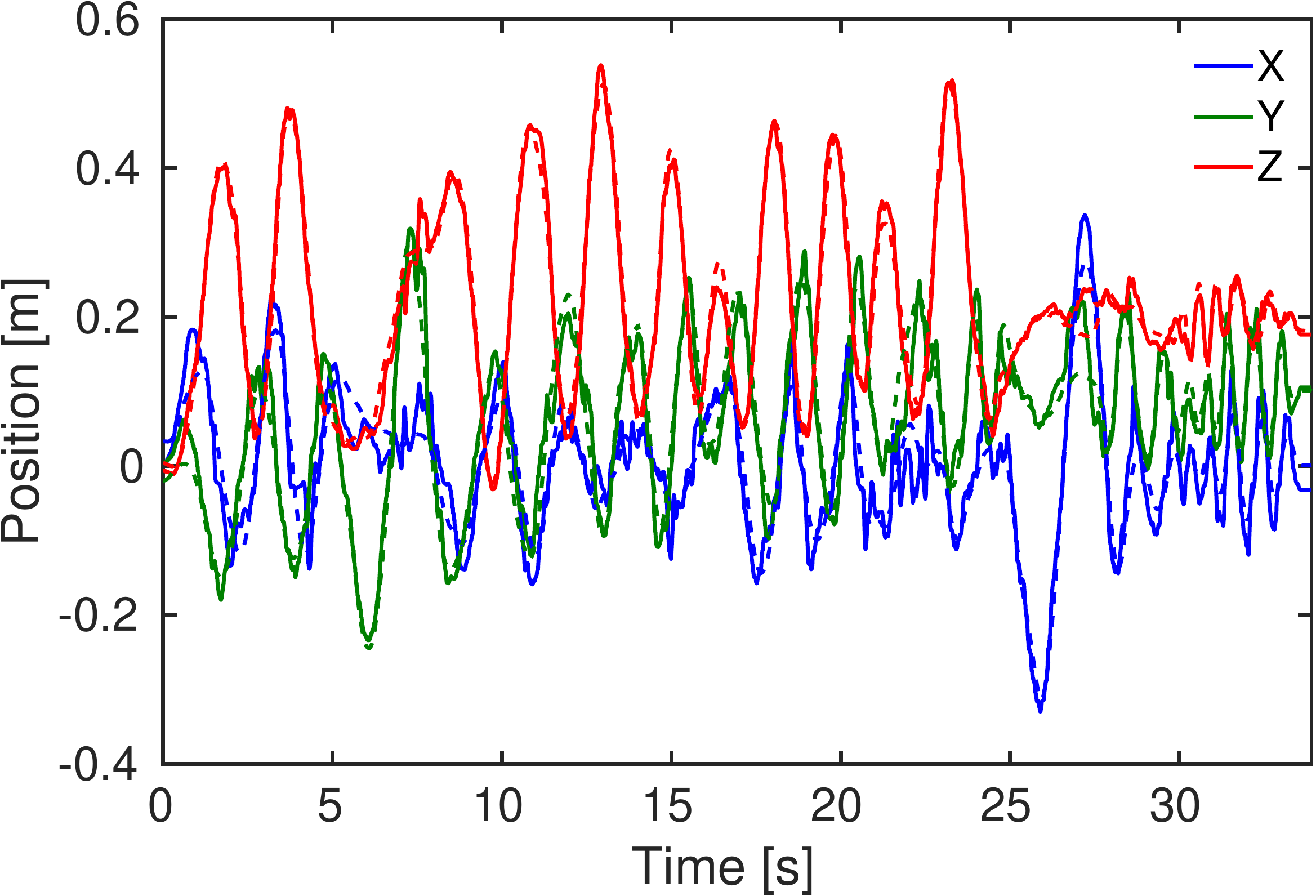}}\;\;
\raisebox{-0.5\height}{\includegraphics[width=0.34\linewidth]{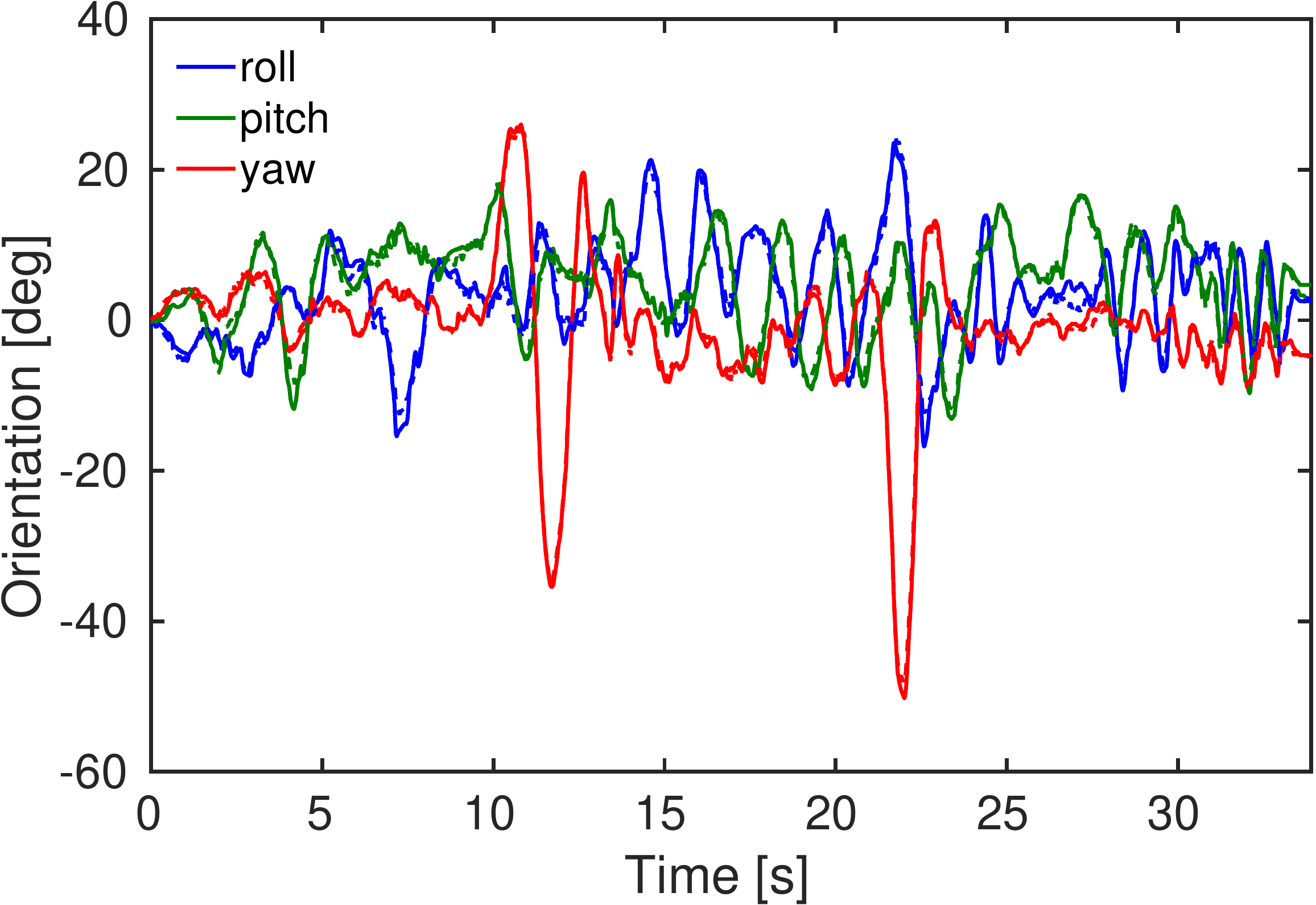}}\\[1ex]
\raisebox{-0.4\height}{\includegraphics[width=0.26\linewidth]{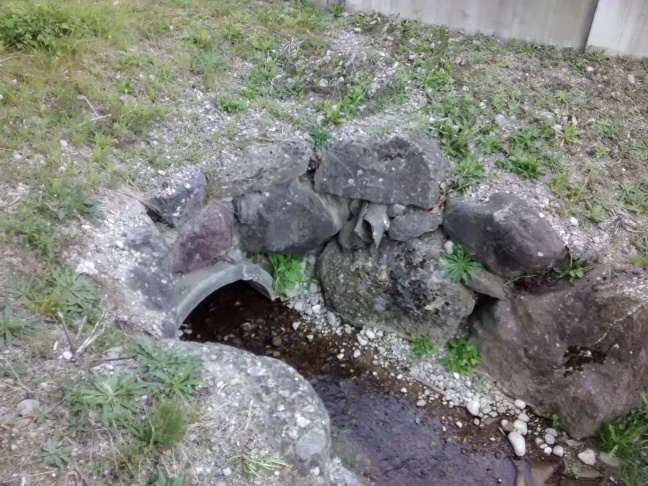}}\;\;\;
\raisebox{-0.5\height}{\includegraphics[width=0.34\linewidth]{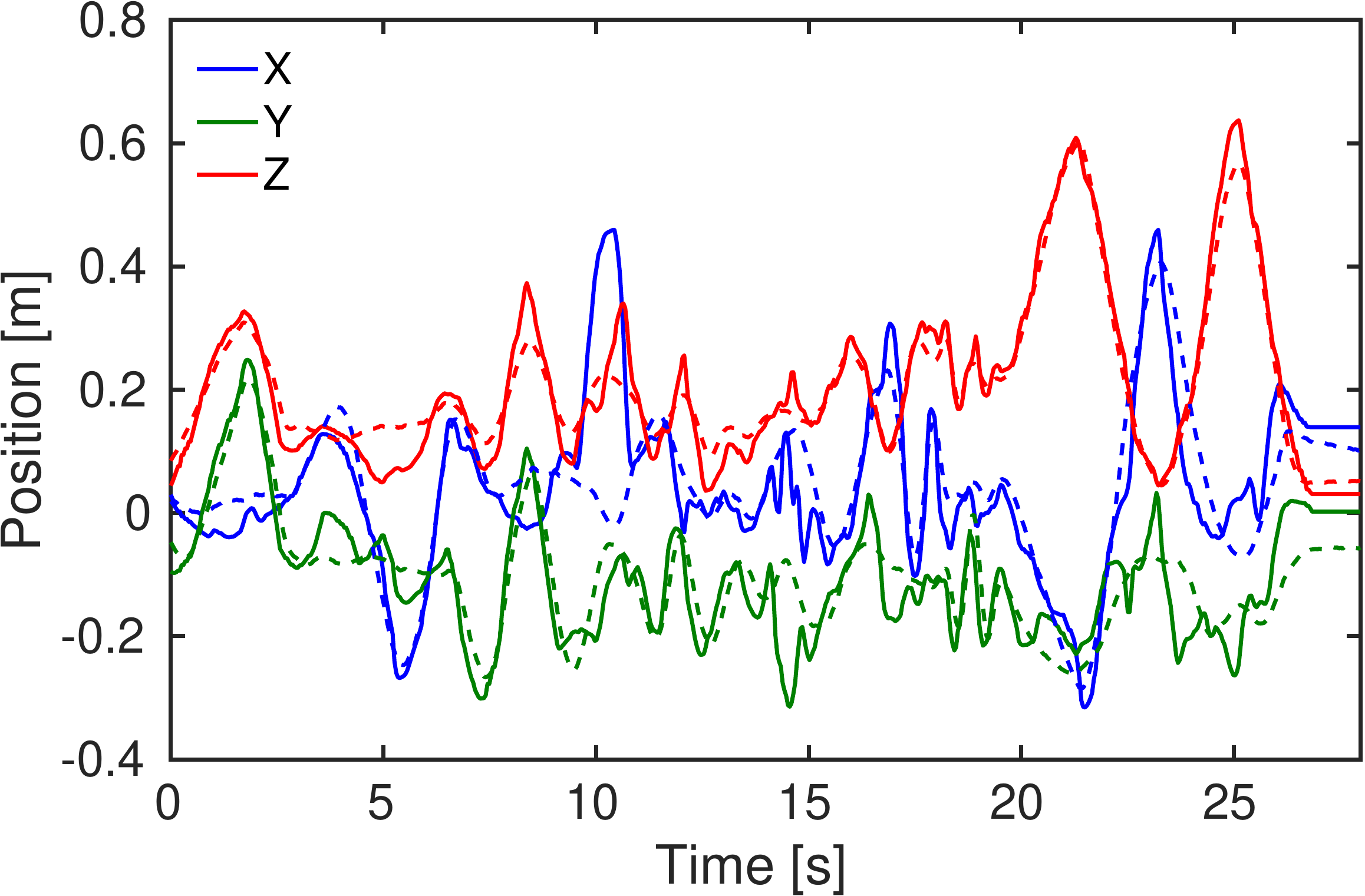}}\;\;
\raisebox{-0.5\height}{\includegraphics[width=0.34\linewidth]{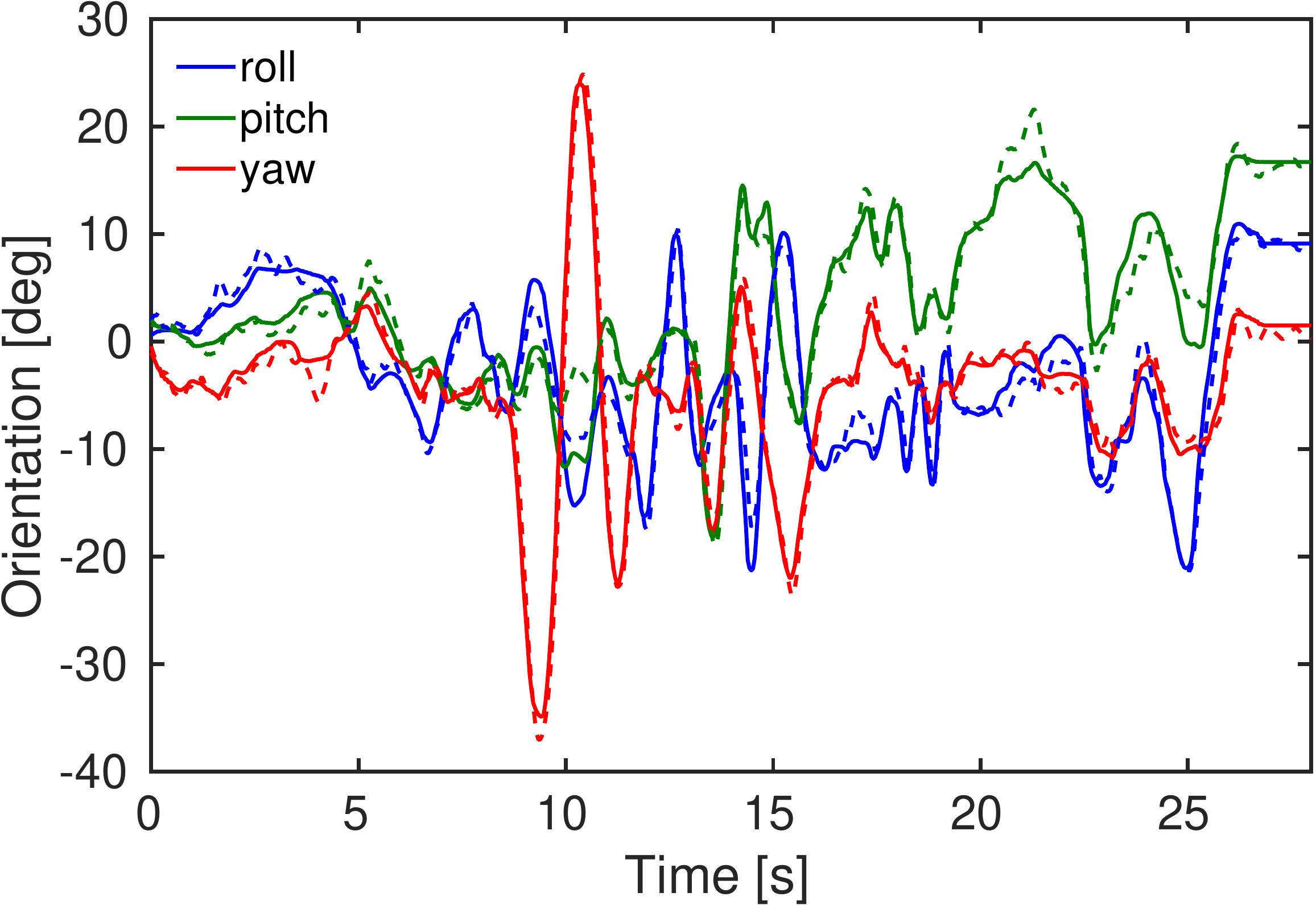}}\\[1ex]
\raisebox{-0.4\height}{\includegraphics[width=0.26\linewidth]{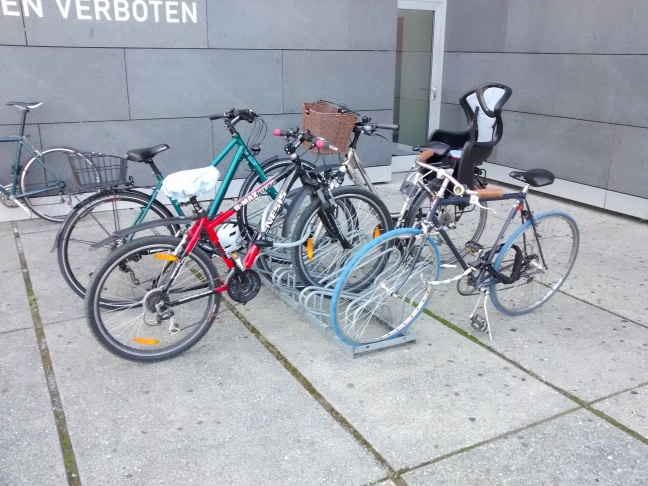}}\;\;
\raisebox{-0.5\height}{\includegraphics[width=0.34\linewidth]{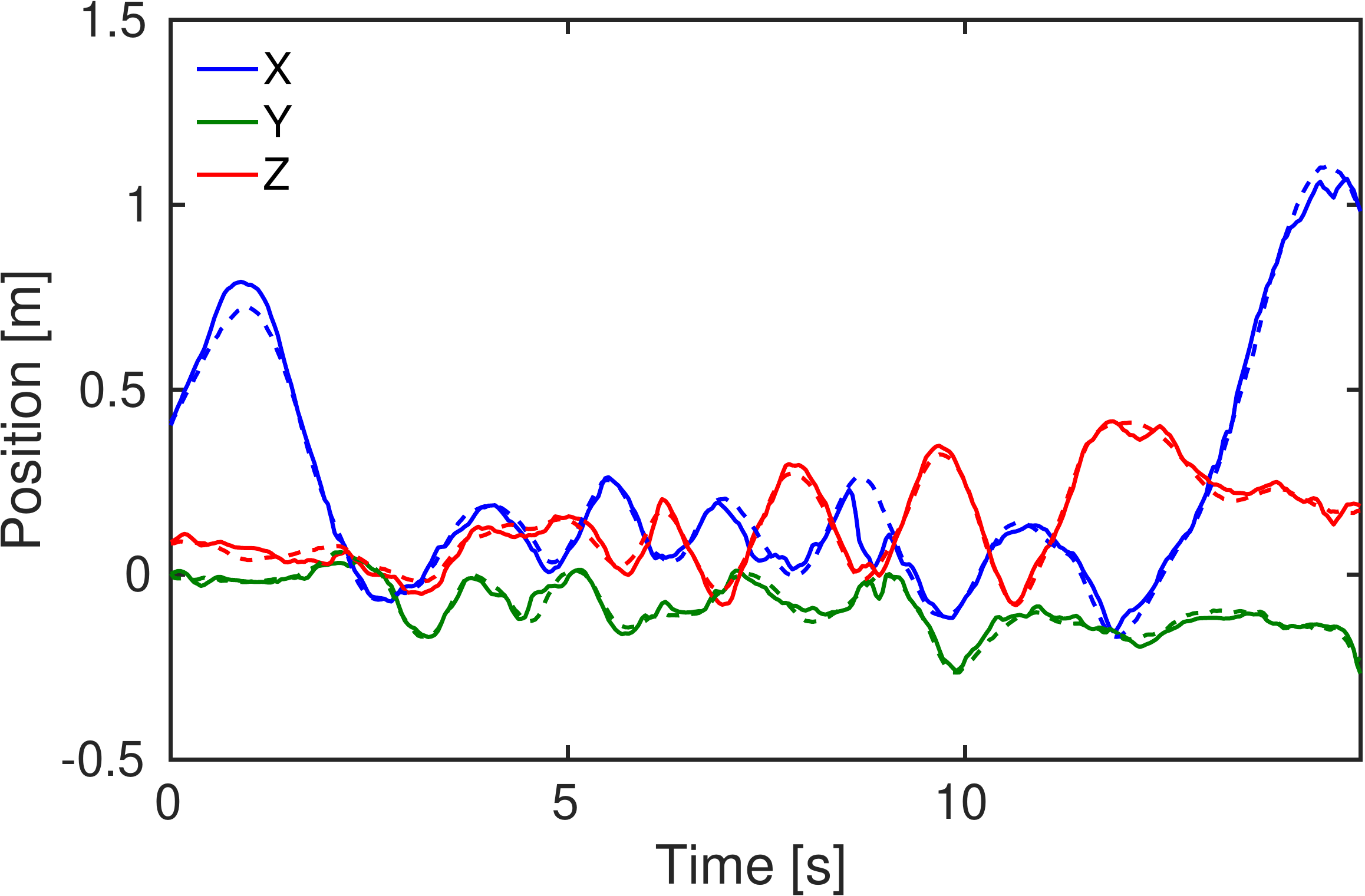}}\;\;
\raisebox{-0.5\height}{\includegraphics[width=0.34\linewidth]{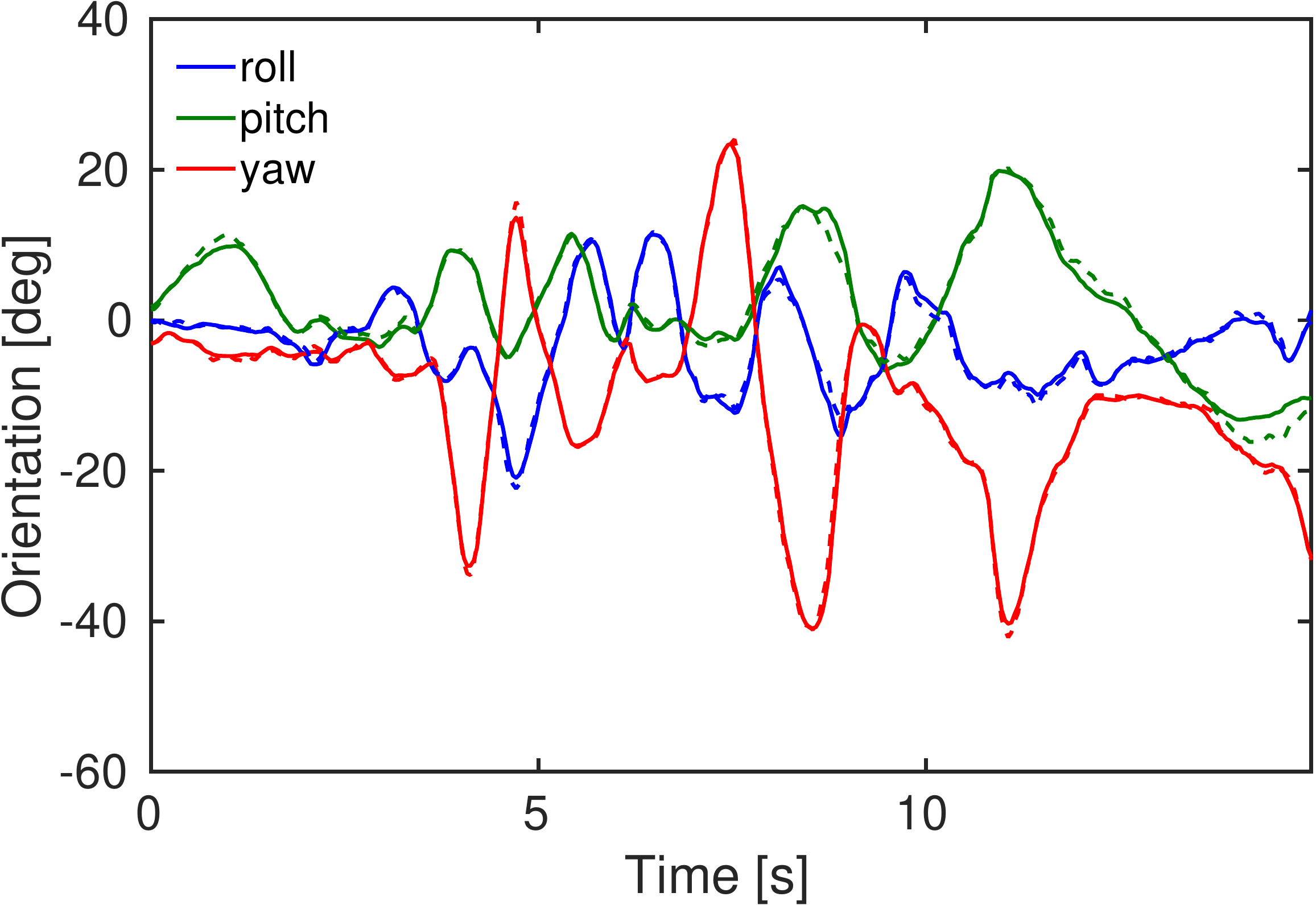}}
   \caption{
   Experiments on scenes with significant depth variation and occlusions.
   Scene impressions (1st column): {\footnotesize\texttt{boxes}}, {\footnotesize\texttt{pipe}}, and {\footnotesize\texttt{bicycles}}.
   Estimated position (2nd column, in meters) and orientation (3rd column, in degrees) from our event-based algorithm (solid line) compared with ground truth (dashed line).
   The mean scene depths are \SI{1.8}{\meter}, \SI{2.7}{\meter}, and \SI{2.3}{\meter}, respectively.}
\label{fig:PlotsParallaxScenes}
\end{figure*}

In the following set of experiments, we also assessed the accuracy of our method on scenes with large depth variation and, therefore larger parallax than in previous experiments.
We recorded seven sequences with ground truth from a motion-capture system 
of a scene consisting of a set of textured {\footnotesize\texttt{boxes}} (Fig.~\ref{fig:PlotsParallaxScenes}, top row). 
We also recorded two outdoor sequences: {\footnotesize\texttt{pipe}} and {\footnotesize\texttt{bicycles}} (middle and bottom rows of Fig.~\ref{fig:PlotsParallaxScenes}). 
The {\footnotesize\texttt{pipe}} sequence depicts a creek going through a pipe, surrounded by rocks and grass; 
the {\footnotesize\texttt{bicycle}} sequence depicts some parked bicycles next to a building; 
both outdoor scenes present some occlusions. 
All sequences exhibit significant translational and rotational motion.

\begin{figure}[t!]
\centering
\raisebox{-0.5\height}{\includegraphics[width=0.48\columnwidth]{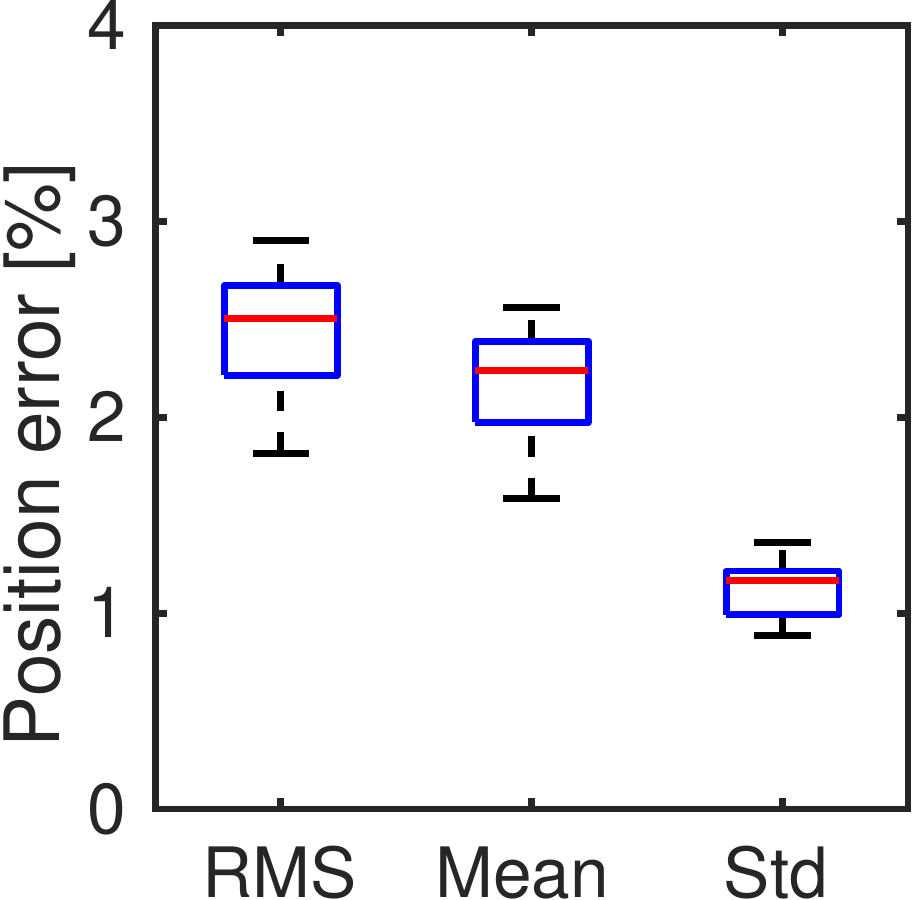}}\;\;\;\raisebox{-0.5\height}{\includegraphics[width=0.48\columnwidth]{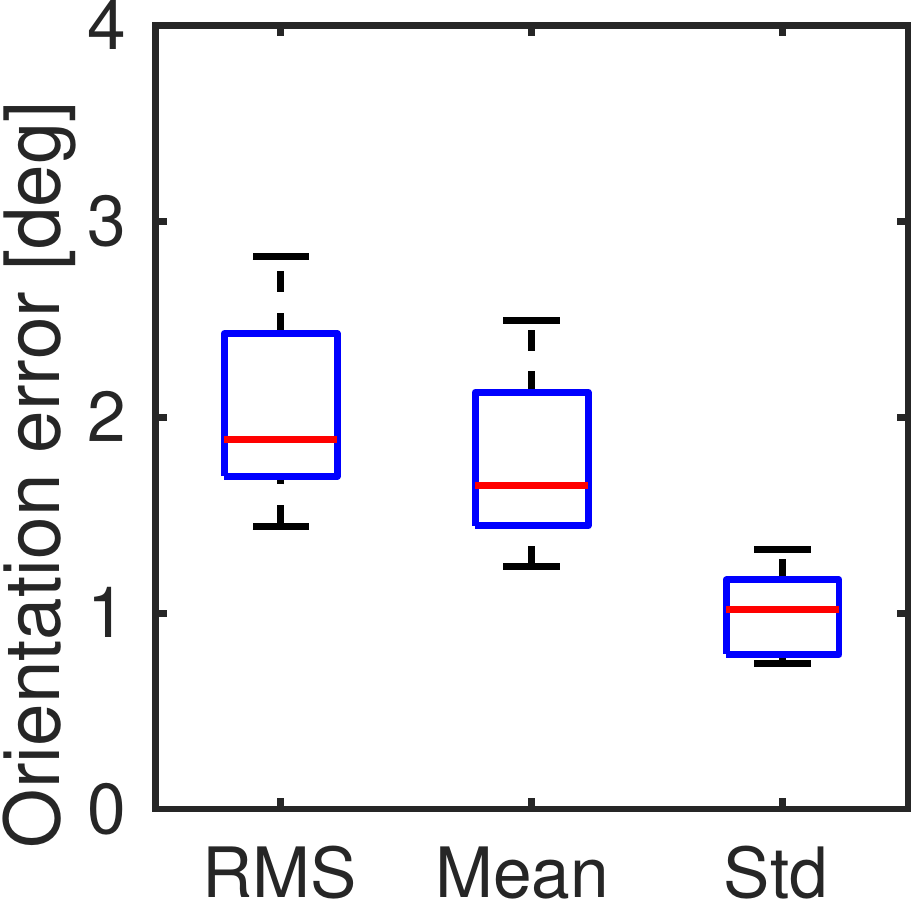}}
   \caption{Error in position (relative to a mean scene depth of \SI{1.9}{\meter}) and orientation (in degrees) of the trajectories recovered by our method for \emph{all} {\footnotesize\texttt{boxes}} sequences (ground truth is given by a motion-capture system).
   We provide box plots of the root-mean-square (RMS) errors, the mean errors and the standard deviation (Std) of the errors.}
\label{fig:BoxPlotsOptitrackBoxes}
\end{figure}

Fig.~\ref{fig:BoxPlotsOptitrackBoxes} summarizes the position and orientation error statistics of our algorithm on the {\footnotesize\texttt{boxes}} sequences (compared with ground truth from the motion-capture system).
The position error is given relative to the mean scene depth, which is \SI{1.9}{\meter}.
As it is observed, the errors are very similar to those in Fig.~\ref{fig:BoxPlotsOptitrack}, meaning that our pose tracking method can handle arbitrary 3D scenes, i.e., not necessarily nearly planar.

Table~\ref{tab:PoseErrorsParallax} reports the numerical values of the trajectory errors in both indoors and outdoor sequences.
The row corresponding to the {\footnotesize\texttt{boxes}} sequences is the average of the errors in the seven indoor sequences (Fig.~\ref{fig:BoxPlotsOptitrackBoxes}).
For the position error, the mean scene depths of the {\footnotesize\texttt{pipe}} and {\footnotesize\texttt{bicycles}} sequences are \SI{2.7}{\meter} and \SI{2.2}{\meter}, respectively. 
The mean RMS errors in position and orientation are in the range \SIrange{2.5}{4.0}{\percent} of the mean scene depth and \SIrange{1.4}{2.9}{\degree}, respectively, 
which are in agreement with the values in Table~\ref{tab:PoseErrors} for the scenes with mean depths smaller than \SI{3}{\meter}.
It is remarkable that the method is able to track despite some lack of texture (as in the {\footnotesize\texttt{pipe}} sequence, where there are only few strong edges), and in the presence of occlusions, which are more evident in the {\footnotesize\texttt{bicycles}} sequence.

\begin{table}[t!]
  \centering
  \caption{Error measurements of the sequences in Fig.~\ref{fig:PlotsParallaxScenes}.
  Translation errors are relative (\ie scaled by the mean scene depth).}
  \label{tab:PoseErrorsParallax}
  \begin{tabular}{l|SSS|SSS}
    \toprule
    & \multicolumn{3}{c|}{Position error [\si{\percent}]} & 
    \multicolumn{3}{c}{Orientation error [\si{\degree}]} \\
    & RMS & $\mu$ & $\sigma$ & RMS & $\mu$ & $\sigma$ \\
    \midrule
    {\footnotesize\texttt{boxes}} & 2.50 & 2.23 & 1.17 & 1.88 & 1.65 & 1.02 \\
    {\footnotesize\texttt{pipe}} & 4.04 & 3.04 & 2.66 & 2.90 & 2.37 & 1.67 \\
    {\footnotesize\texttt{bicycles}} & 2.14 & 1.724 & 1.27 & 1.46 & 1.19 & 0.84\\
    \bottomrule
  \end{tabular}
\end{table}

\subsection{Computational Effort}
We measured the computational cost of our algorithm on a single core of an Intel(R) i7 processor at \SI{2.60}{\giga\hertz}.
The processing time per event is $\SI{32}{\micro\second}$, resulting in a processing event rate of 31.000 events per second. 
Depending on the texture of the scene and the speed of motion, the data rate produced by an event camera ranges from tens of thousands (moderate motion) to over a million events per second (high-speed motion). 
However, notice that our implementation is not optimal; many computations can indeed still be optimized, cached, and parallelized to increase the runtime performance of the algorithm.

\section{Conclusion}
\label{sec:Conclusion}
We have presented an approach to track the \mbox{6-DOF} pose of an arbitrarily moving event camera from an existing photometric depth map in natural scenes.
Our approach follows a Bayesian filtering methodology: 
the sensor model is given by a mixture-model likelihood that takes into account both the event-generation process and the presence of noise and outliers;
the posterior distribution of the system state is approximated according to the relative-entropy criterion using distributions in the exponential family and conjugate priors.
This yields a robust EKF-like filter that provides pose updates for every incoming event, at microsecond time resolution.

We have compared our method against ground truth from a motion capture system and a state-of-the-art frame-based pose-tracking pipeline.
The experiments revealed that the proposed method accurately tracks the pose of the event-based camera, both in indoor and outdoor experiments in scenes with significant depth variations, and under motions with excitations in all 6-DOFs.

\ifCLASSOPTIONcaptionsoff
  \newpage
\fi


\bibliographystyle{IEEEtran}
\bibliography{all}



\begin{IEEEbiography}[{\includegraphics[width=1in,height=1.25in,clip,keepaspectratio]{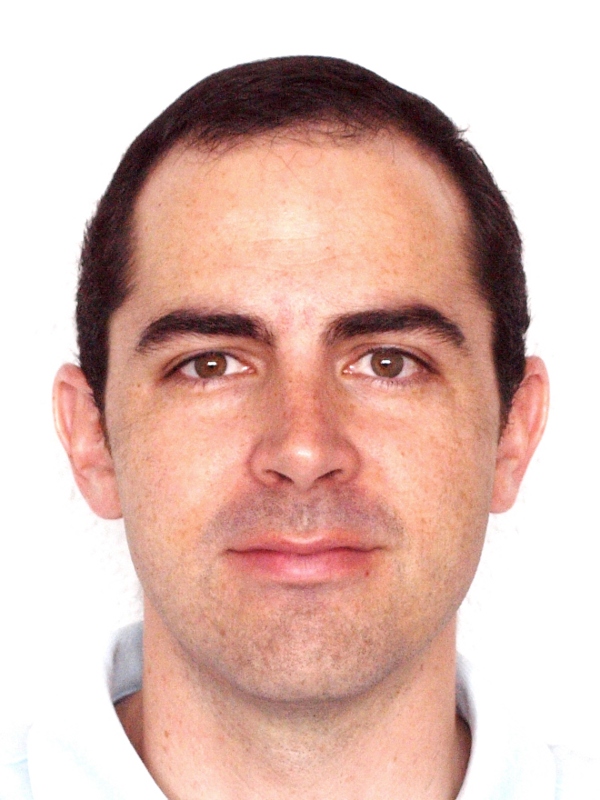}}]{Guillermo Gallego}
received the Ph.D. degree in electrical and computer engineering from the Georgia Institute of Technology, Atlanta, GA, USA, in 2011.
He received the Ingeniero de Telecomunicaci\'on degree (five-year engineering program) from the Universidad Polit\'ecnica de Madrid (UPM), Madrid, Spain, in  2004, 
the M.S. degree in mathematical engineering (Mag\'ister en Ingenier\'ia Matem\'atica) from the Universidad Complutense de Madrid, Madrid, in 2005, 
and the M.S. degrees in electrical and computer engineering and mathematics from the Georgia Institute of Technology, Atlanta, GA, USA, in 2007 and 2009, respectively.
From 2011 to 2014, he was a Marie Curie Post-Doctoral Researcher with the UPM. 
Since 2014, he has been with the Robotics and Perception Group, University of Zurich, Zurich, Switzerland. 
His current research interests include computer vision, signal processing, robotics, geometry, optimization, and ocean engineering.
Dr. Gallego was a recipient of the Fulbright Scholarship to pursue graduate studies at the Georgia Institute of Technology in 2005.
He is a recipient of the Misha Mahowald Prize for Neuromorphic Engineering (2017).
\end{IEEEbiography}

\begin{IEEEbiography}[{\includegraphics[width=1in,height=1.25in,clip,keepaspectratio]{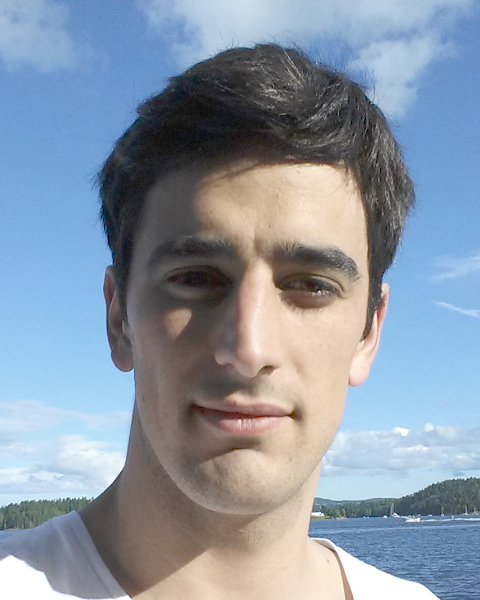}}]{Jon E.A. Lund} is a flight control and video engineer at Prox Dynamics/FLIR UAS in Oslo, Norway. 
He received his M.Sc. (2015) in Neural Systems and Computation from ETH and University of Zurich, Switzerland. 
Before that, he earned a B.Sc. (2013) in Physics at the University of Oslo, Norway.
\end{IEEEbiography}

\begin{IEEEbiography}[{\includegraphics[width=1in,height=1.25in,clip,keepaspectratio]{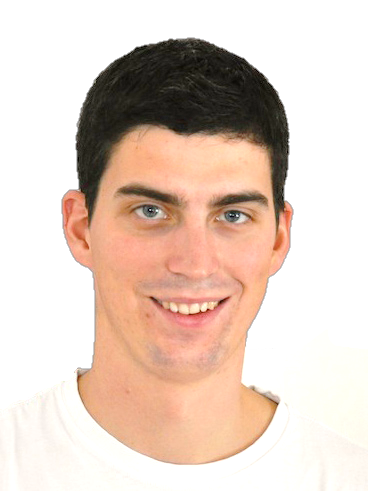}}]{Elias Mueggler} 
received his Ph.D. degree from the University of Zurich, Switzerland in 2017, where he was working at the Robotics and Perception Group, lead by Prof. Davide Scaramuzza, in the topics of event-based vision for high-speed robotics and air-ground robot collaboration.
He received B.Sc. (2010) and M.Sc. (2012) degrees in Mechanical Engineering from ETH Zurich, Switzerland.
He is a recipient of the KUKA Innovation Award (2014), the Qualcomm Innovation Fellowship (2016) and the Misha Mahowald Prize for Neuromorphic Engineering (2017).
He has been a visiting researcher with Prof. John Leonard (Massachusetts Institute of Technology) and Dr. Chiara Bartolozzi (Istituto Italiano di Tecnologia).
His research interests include computer vision and robotics.
\end{IEEEbiography}

\vspace{-1ex}
\begin{IEEEbiography}[{\includegraphics[width=1in,height=1.25in,clip,keepaspectratio]{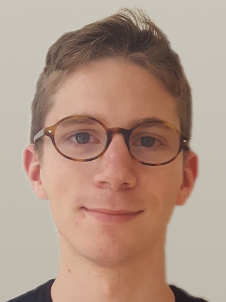}}]{Henri Rebecq} is a Ph.D. student in the Robotics and Perception Group at the University of Zurich, where he is working on on event-based vision for robotics.
In 2014, he received an M.Sc.Eng. degree from T\'el\'ecom ParisTech, 
and an M.Sc. degree from Ecole Normale Sup\'erieure de Cachan, both located in Paris, France.
Prior to pursuing graduate studies, he worked as a research and software engineer at VideoStitch, Paris, France.
His research interests include omnidirectional vision, visual odometry, 3D reconstruction and SLAM.
He is a recipient of the Misha Mahowald Prize for Neuromorphic Engineering (2017).
\end{IEEEbiography}


\vspace{-1ex}
\begin{IEEEbiography}[{\includegraphics[width=1in,height=1.25in,clip,keepaspectratio]{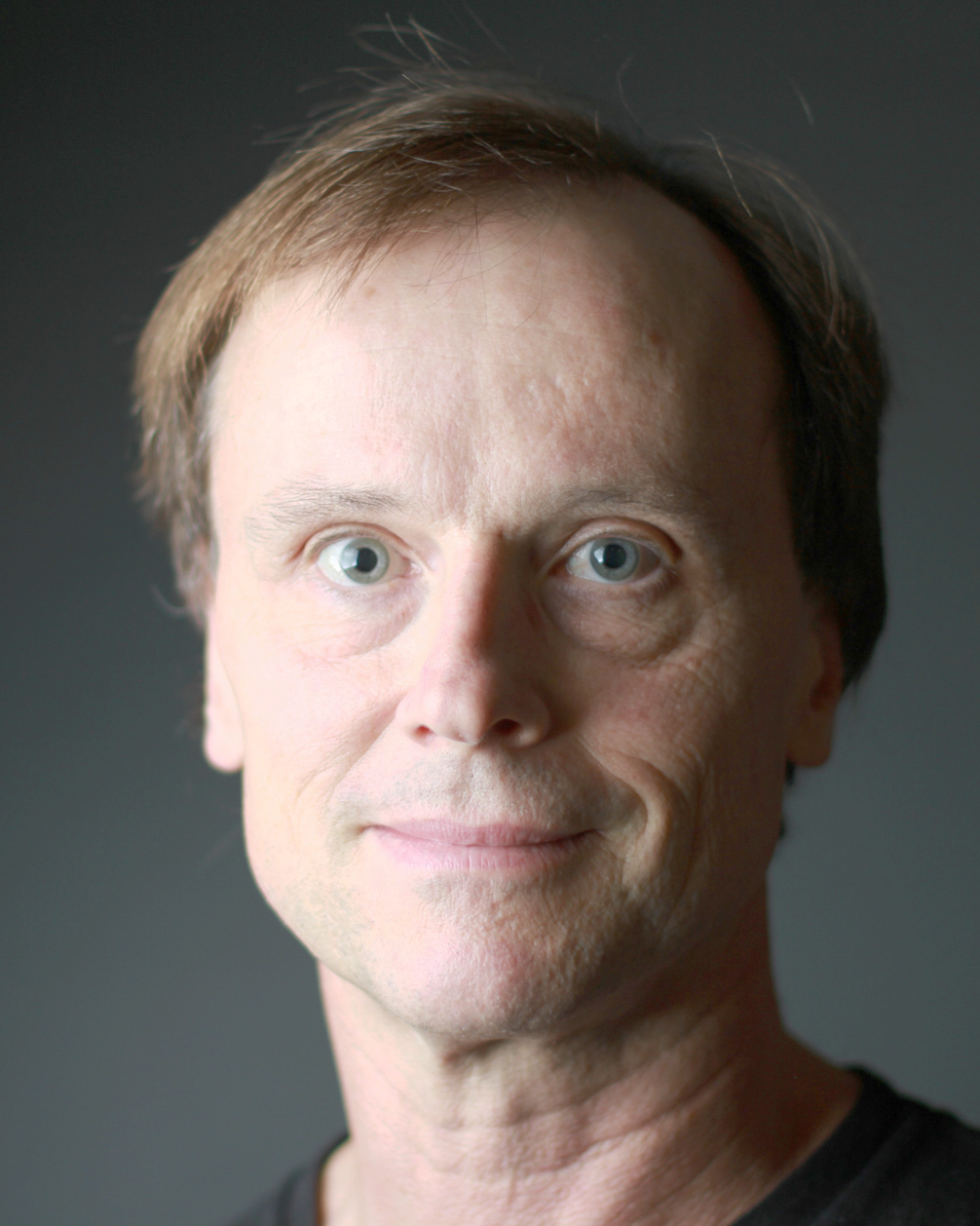}}]{Tobi Delbruck}
(M'99--SM'06--F'13) received the B.Sc. degree in physics and applied mathematics from the University of California, San Diego, CA, USA, and the Ph.D. degree from the California Institute of Technology, Pasadena, CA, USA, in 1986 and 1993, respectively.
He has been a Professor of Physics with the Institute of Neuroinformatics, University of Zurich and ETH Zurich, Switzerland, since 1998. 
His group focuses on neuromorphic sensory processing. 
He worked on electronic imaging at Arithmos, Synaptics, National Semiconductor, and Foveon.
Dr. Delbruck has co-organized the Telluride Neuromorphic Cognition Engineering summer workshop and the live demonstration sessions at International Symposium on Circuits and Systems.  
He is also co-founder of iniLabs and Insightness.  
He was the Chair of the IEEE CAS Sensory Systems Technical Committee, is current  Secretary of the IEEE Swiss CAS/ED Society, and an Associate Editor of the IEEE Transactions on Biomedical Circuits and Systems. 
He has received 9 IEEE awards.
\end{IEEEbiography}

\vspace{-1ex}
\begin{IEEEbiography}[{\includegraphics[width=1in,height=1.25in,clip,keepaspectratio]{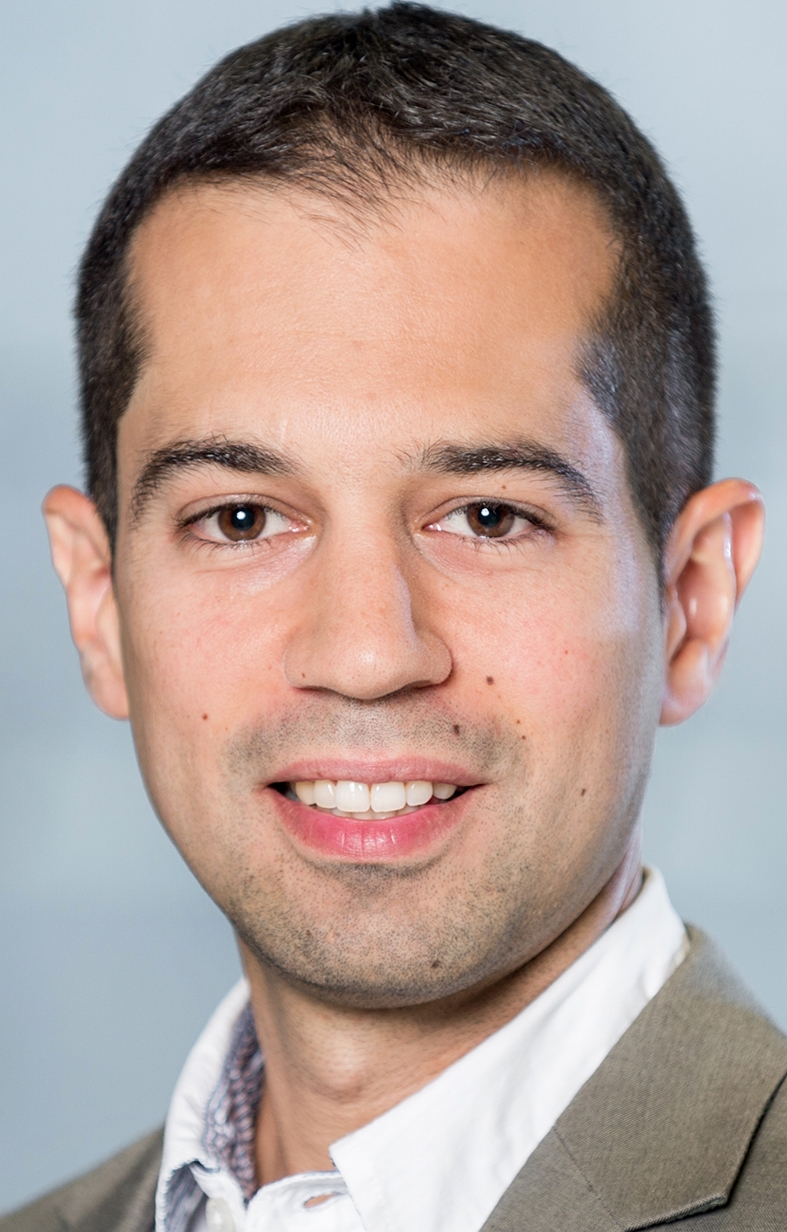}}]{Davide Scaramuzza}
Davide Scaramuzza (born in 1980, Italian) is Associate Professor of Robotics and Perception at both departments of Informatics (University of Zurich) and Neuroinformatics (University of Zurich and ETH Zurich), where he does research at the intersection of robotics, computer vision, and neuroscience. He did his PhD in robotics and computer vision at ETH Zurich and a postdoc at the University of Pennsylvania. From 2009 to 2012, he led the European project “sFly”, which introduced the PX4 autopilot and pioneered visual-SLAM--based autonomous navigation of micro drones. For his research contributions, he was awarded the Misha Mahowald Neuromorphic Engineering Award, the IEEE Robotics and Automation Society Early Career Award, the SNSF-ERC Starting Grant, a Google Research Award, the European Young Research Award, and several conference paper awards. He coauthored the book “Introduction to Autonomous Mobile Robots” (published by MIT Press) and more than 100 papers on robotics and perception.
\end{IEEEbiography}




\appendices
\section{Rewriting the Likelihood Function}
\label{sec:LikelihoodAsSumOfExp}
A distribution in the exponential family can be written as
\begin{equation}
\label{eq:DefExpFamilyDistrib}
p(x;\natparam) = \basemeas(x)\exp\left(\natparam \cdot \suffStat(x) - A(\natparam)\right),
\end{equation}
where $\natparam$ are the natural parameters, $T(x)$ are the sufficient statistics of $x$, $A(\natparam)$ is the log-normalizer, and $h(x)$ is the base measure.

The likelihood~
\textcolor{magenta}{(15)} can be rewritten as:
\begin{align}
\label{eq:likelihood2c} p(\obs_k|\state_k) = & \frac1{\sqrt{2\pi}}\exp(\log(\inlierProb)-\log(\stdThr)\\
\nonumber &-\frac1{2}\left[\JacMeas^{i}\JacMeas^{i}\frac{\tilde{\state}_k^{i}\tilde{\state}_k^{j}}{\varThr}+2\Mo\JacMeas^{i}\frac{\tilde{\state}_k^{i}}{\varThr} + \frac{\Mo^{2}}{\varThr}\right]\\
\nonumber &+\exp(\log((1-\inlierProb)/(M_{\max}-M_{\min}))),
\end{align}
where we use the Einstein summation convention for the indices of $\JacMeas = (\JacMeas^{i})$ and $\tilde{\state}_k = (\tilde{\state}_k^{i})$. 
Collecting the sufficient statistics into
\[
\suffStat(\state_k) = \left[\frac{\tilde{\state}_k^{i}\tilde{\state}_k^{j}}{\varThr},\frac{\tilde{\state}_k^{i}}{\varThr},\frac1{\varThr},\log(\stdThr),\log(\inlierProb),\log(1-\inlierProb)\right],
\]
the likelihood can be conveniently rewritten as a sum of two exponential families~
\textcolor{magenta}{(16)}, $j=1,2$, with $\basemeas(s)=1$,
\begin{align}
\label{eq:likelihood_statistics1} \natparam_{o,1} & = \left[-\frac1{2}\JacMeas^{i}\JacMeas^{j},-\Mo\JacMeas^{i},-\frac1{2}\Mo^{2},-1,1,0\right]\\
\label{eq:likelihood_statistics2} \natparam_{o,2} & = \left[0_{ij},0_{i},0,0,1\right] \\
\label{eq:likelihood_statistics3} A_{o,1} & = \log\sqrt{2\pi}\\
\label{eq:likelihood_statistics4} A_{o,2} & = -\log(M_{\max} - M_{\min}).
\end{align}

\end{document}